\documentclass[twoside, 11pt, pdflatex]{article}

\usepackage[preprint]{jmlr2e}

\usepackage{algorithm}
\usepackage{algorithmic}
\usepackage{mathtools}
\usepackage{subfigure}
\usepackage{url}

\newcommand*{\eqspace}{\;\;}

\jmlrheading{1}{2020}{1-1}{4/00}{10/00}{meila00a}{Shunki Kyoya and Kenji Yamanishi}

\ShortHeadings{Mixture Complexity and Gradual Clustering Change}{Kyoya and Yamanishi}
\firstpageno{1}

\begin{document}

\title{Mixture Complexity and Its Application to Gradual Clustering Change Detection}

\author{\name Shunki Kyoya \email kyoya\_shunki@mist.i.u-tokyo.ac.jp \\
  \name Kenji Yamanishi \email yamanishi@mist.i.u-tokyo.ac.jp \\
  \addr Graduate School of Information Science and Technology \\
  The University of Tokyo \\
  7-3-1 Hongo, Bunkyo-ku, Tokyo, 113-8656, Japan
}

\editor{}

\maketitle

\begin{abstract}%   <- trailing '%' for backward compatibility of .sty file
  In model-based clustering using finite mixture models,
  it is a significant challenge to determine the number of clusters (cluster size).
  It used to be equal to the number of mixture components (mixture size);
  however, this may not be valid in the presence of overlaps or weight biases.
  In this study, we propose to continuously measure the cluster size in a mixture model
  by a new concept called mixture complexity (MC).
  It is formally defined from the viewpoint of information theory
  and can be seen as a natural extension of the cluster size considering overlap and weight bias.
  Subsequently, we apply MC to the issue of gradual clustering change detection.
  Conventionally, clustering changes has been considered to be abrupt,
  induced by the changes in the mixture size or cluster size.
  Meanwhile, we consider the clustering changes to be gradual in terms of MC;
  it has the benefits of finding the changes earlier and discerning the significant and insignificant changes.
  We further demonstrate that the MC can be decomposed according to the hierarchical structures of
  the mixture models; it helps us to analyze the detail of substructures.
\end{abstract}

\begin{keywords}
  Finite mixture model,
  Clustering,
  Change detection,
  Gradual change,
  Information theory
\end{keywords}

\section{Introduction}

As an introduction, we first state our interest
and roughly introduce the concept of mixture complexity.
Then, we discuss its related work and significance.

\subsection{Motivation}

Finite mixture models are widely used for model-based clustering
(for overviews and references see \citet{McLachlanPeel2000, FraleyRaftery1998}).
In this field, it is a classical issue to determine the number of components.
It has the following two meanings:
the number of elements to represent the density distribution
and the number of clusters to group the data
(referred to as \emph{mixture size} and \emph{cluster size}, respectively).
In this study, we consider the problem of estimating the cluster size when the mixture size is given.
The cluster size used to be equal to the mixture size;
however, it may not be valid when the components have overlaps or weight biases.
Therefore, we need to reconsider the definitions and meanings of the cluster size.

\begin{figure}[htbp]
  \centering
  \subfigure[]{
    \includegraphics[width=0.3\columnwidth]{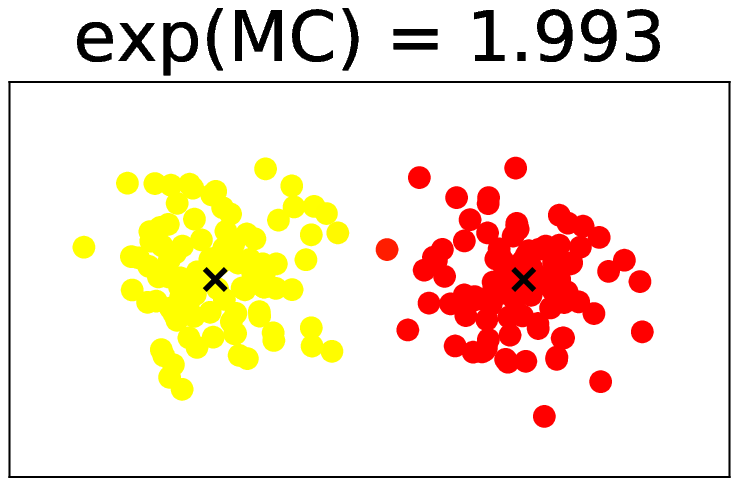}
  }
  \subfigure[]{
    \includegraphics[width=0.3\columnwidth]{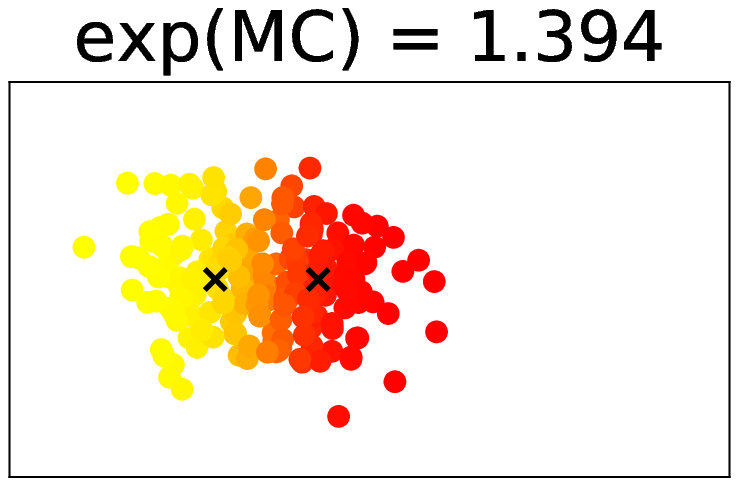}
  }
  \subfigure[]{
    \includegraphics[width=0.3\columnwidth]{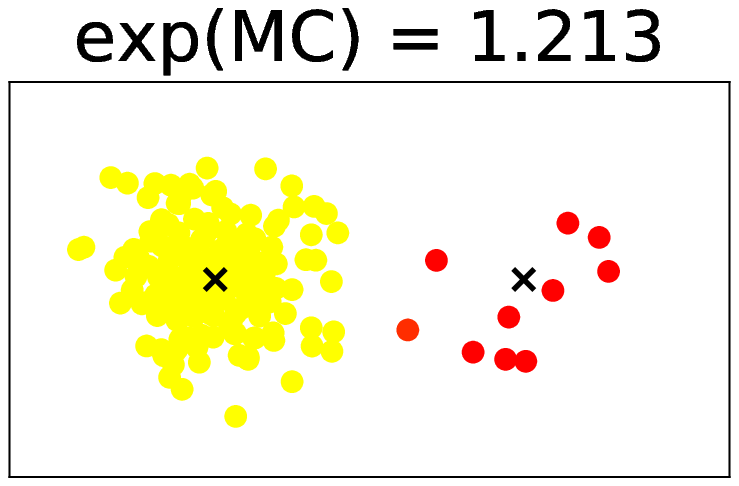}
  }
  \caption{
    Examples of MC with Gaussian mixture models with a mixture size of two.
    \label{Fig: introduction}
  }
\end{figure}

For instance, let us observe three cases of the Gaussian mixture model,
as shown in Figure \ref{Fig: introduction}.
Although the mixture size is two in any case, the situations are different.
In case (a), the two components are distinct from each other and their weights are not biased;
then, there seems to be no problem to believe that the cluster size is two as well.
Meanwhile, in case (b), although their weights are not biased,
the two components are very close to each other;
then, as proposed in \citet{Hennig2010},
we may need to regard them as one cluster by merging them.
In case (c), although the two components are distinct from each other,
their weights are biased;
then, as proposed in \citet{Jiang+2001} and \citet{He+2003},
we may need to regard the small component as outliers rather than a cluster.
Overall, in cases (b) and (c), it may be more difficult to say
that the cluster size is exactly two than that in case (a).
This observation gives rise to the problem of
formally defining the complexity of clustering structures
that reflects the overlaps and weight biases.

This paper introduces a novel concept of \emph{mixture complexity} (MC) to resolve this problem.
It is related to the logarithm of the cluster size.
For example, the exponentials of the MC are 2.00, 1.39, 1.21 for cases (a), (b), and (c), respectively.
In other words, given the mixture size,
MC estimates the cluster size continuously rather than discretely.

There are two reasons for the need of MC.
First, it theoretically evaluates the cluster size in the finite mixture model
considering the overlap and imbalance between the components.
Although their impacts on the cluster size have been discussed independently,
we present a unified framework to interpret the cluster size by a continuous index.
It presents a new perspective on model-based clustering and
can be practically applied to cluster merging or clustering-based outlier detection.
The second is the application of MC to the issue of gradual clustering change detection.
Conventionally, clustering changes have been considered to be abrupt,
induced by the changes in the mixture size or cluster size.
In reality, however, there are cases where mechanisms for generating data change gradually
(or incrementally in the context of concept drifts \citep{Gama+2014}).
We thereby present a new methodology for tracking such changes by observing MC's changes.

We further show that MC can be used to quantify the cluster size in hierarchical mixture models.
We demonstrate that the MC of a hierarchical mixture model can be decomposed
into the sum of MCs for local mixture models.
It enables us to evaluate the complexity of the substructures as well as the entire structure.

\subsection{Related Work}

The issue of determining the best mixture size or cluster size
(often referred to as model selection) has extensively been studied.
For example, AIC \citep{Akaike1974}, BIC \citep{Schwarz1978}, and  MDL \citep{Rissanen1978}
have been used to select the mixture size;
ICL \citep{Biernacki+2000} and MDL-based clustering criterion
\citep{Kontkanen+2005, HiraiYamanishi2013, HiraiYamanishi2019}
have been invented to select the cluster size.
See also a recent review by \citet{McLachlanRathnayake2014}
focusing on the number of components in a Gaussian mixture model.

Differences between the mixture size and cluster size have also been widely discussed.
For example, \citet{McLachlanPeel2000} pointed out that there were cases that
Gaussian mixture models with more than one mixture sizes were needed to describe one skewed cluster;
\citet{Biernacki+2000} argued that in many situations, the mixture size estimated by BIC
was too large to regard it as the cluster size.
The problem of estimating the cluster size under a given mixture size
has also been investigated by \citet{Hennig2010};
he proposed methods to identify the cluster structure by merging heavily overlapped mixture components.
MC differs from his approach in that
it interprets the clustering structure by only measuring the overlap rate
rather than deciding whether to merge based on a certain threshold.

The degree of overlap or closeness between components has been evaluated
using various measures, such as the classification error rate
or the Bhattacharyya distance \citep{Fukunaga1990}.
\citet{Wangsun2004} and \citet{SunWang2011} formulated
the overlap rate of Gaussian distributions from the geometric nature of them.
All of the works above have been limited to the case of two components.
On the other hand, MC considers the overlap between any number of components.

Deciding whether a small component is a cluster or a set of outliers is also a significant matter.
For example, clustering algorithms such as
DBSCAN \citep{Ester+1996} and constrained $k$-means \citep{Bladrey+2000}
avoided generating small components to obtain a better clustering structure.
\citet{Jiang+2001} and \citet{He+2003} associated the small components with outlier detection problems.
MC evaluates the small components by continuously measuring the impacts on the cluster size.

Some other notions have been proposed to quantify the clustering structure.
\citet{Rusch+2018} evaluated the crowdedness of the data under the concept of ``clusteredness.''
However, its relations to the cluster size are indirect.
Recently, descriptive dimensionality (Ddim) \citep{Yamanishi2019}
was proposed to define the model dimensionality continuously.
It can be implemented to estimate the clustering structure under the assumption of model fusion,
that is, models with different number of components are probabilistically mixed.
MC differs from Ddim because it evaluates the overlap and weight bias in the single model
without the model fusion.

Clustering under the data stream has been discussed with various objectives
\citep{Guha+2000, SongWang2005, Chakrabatri+2006}.
We consider the problem of detecting changes in the cluster structure;
Dynamic model selection (DMS)
\citep{YamanisniMaruyama2005, YamanishiMaruyama2007, HiraiYamanishi2012}
addressed this problem by observing the changes in the models
(corresponding to mixture size or cluster size in this paper).
Because the models are valued discretely, the detected changes have been considered to be abrupt.
Refer also to the notions of tracking best experts \citep{Herbser+1998},
evolution graph \citep{Ntoutsi+2012}, and switching distributions \citep{vanEreven+2012},
which are similar to DMS.

Furthermore, the issues of gradual changes have been discussed
to investigate the transition periods for absolute changes.
The MDL change statistics \citep{YamanishiMiyaguchi2016} was proposed
to measure the degree of gradual changes.
The notions of structural entropy \citep{HiraiYamanishi2018} and graph entropy \citep{Ohsawa2018}
were proposed to measure the degree of model uncertainty in the changes.
This study quantifies the degree of gradual changes
by the fluctuations in MC and presents a new methodology to detect them.

\subsection{Significance and novelty}
The significance and novelty of this paper are summarized below.

\paragraph{Mixture complexity for finite mixture models.}

We introduce a novel concept of MC to continuously measure the cluster size in a mixture model.
It is formally defined from the viewpoint of information theory
and can be interpreted as a natural extension of the cluster size
considering the overlaps and weight biases among the components.
We further demonstrate that MC can be decomposed into a sum of MCs
according to the mixture hierarchies; it helps us in analyzing MC in a decomposed manner.

\paragraph{Applications of MC to gradual clustering change detection.}

We apply MC to the issue of monitoring gradual changes in clustering structures.
We propose methods to monitor changes in MC instead of the mixture size or cluster size.
Because MC takes a real value, it is more suitable for observing gradual changes.
We empirically demonstrate that MC elucidates the clustering structures and their changes more effectively
than the mixture size or cluster size.

\bigskip

The remainder of this paper is organized as follows.
In Section \ref{Sec: MC}, we introduce the concept of MC.
We also present some examples and theoretical properties.
In Section \ref{Sec: decomposition}, we show the decomposition properties of MC.
Section \ref{Sec: application} discusses an application of MC to clustering change detection problems
and Section \ref{Sec: results} describes the experimental results.
Finally, Section \ref{Sec: conclusion} concludes this paper.
Proofs of the propositions and theorems are described in Appendix \ref{Sec: proofs}.
Programs for the experiments are available at 
\url{https://github.com/ShunkiKyoya/MixtureComplexity}.

\section{Mixture Complexity \label{Sec: MC}}

In this section, we formally introduce the mixture complexity and describe its properties
by some examples and theories.

\subsection{Definitions}

Given the data $\{ x_n \}_{n = 1}^N$ and the finite mixture model $f$ that have generated them,
we consider estimating the cluster size of $f$.
The distribution $f$ is written as
\[
  f(x) \coloneqq \sum_{k = 1}^K \pi_k g_k(x) \eqspace ,
\]
where $K$ denotes the mixture size,
$\{ \pi_k \}_{k = 1}^K$ denote the proportions of each component summing up to one,
and $\{ g_k \}_{k = 1}^K$ denote the probability distributions.
The random variable $X$ following the distribution $f$ is called an \emph{observed variable}
because it can be observed as a datum.
We also define the \emph{latent variable} $Z \in \{1, \dots, K \}$
as the index of the component from which the observed variable $X$ originated.
The pair $(X, Z)$ is called a \emph{complete variable}.
The distribution of the latent variable $p(Z)$
and the conditional distribution of the observed variable $p(X | Z)$ can be given by
\[
  p(Z = k) = \pi_k \eqspace , \quad p(X | Z = k) = g_k (X) \eqspace .
\]

To investigate the clustering structures in $f$, we consider the following quantity:
\[
  I(Z; X)
  \coloneqq H(Z) - H(Z | X)
  = \int \mathrm{d}x \sum_{k = 1}^K \pi_k g_k(x)
  \log \left( \frac{g_k(x)}{f(x)} \right) \eqspace ,
\]
where $H(Z)$ and $H(Z | X)$ denote the entropy and conditional entropy, respectively,
of the latent variable $Z$ defined as
\begin{align*}
  H(Z)     & \coloneqq - \sum_{k = 1}^K p(Z = k) \log p(Z = k)
  = - \sum_{k=1}^K \pi_k \log \pi_k \eqspace ,                                 \\
  H(Z | X) & \coloneqq - \int \mathrm{d}x \sum_{k = 1}^K p(X, Z) \log p(Z | X)
  = - \int \mathrm{d}x \sum_{k = 1}^K \pi_k g_k(x)
  \log \left( \frac{\pi_k g_k(x)}{f(x)} \right) \eqspace .
\end{align*}
The quantity $I(Z; X)$ is well-known
as the mutual information between the observed and latent variables;
it is also known as the (generalized) Jensen-Shannon Divergence \citep{Lin1991}.
We can interpret $I(Z; X)$ as the volume of cluster structures as follows.
Because $I(Z; X)$ is a subtraction of the latent variable's entropy
with and without the knowledge of the observed variable,
it represents the amount of information
about the latent variable possessed by the observed data.
Thus, its exponent $\exp (I(X; Z))$ denotes the ``number'' of clusters
distinguished by the observed variable;
it can be interpreted as a continuous extension of the cluster size.
For more information about entropy and mutual information,
see \citet{CoverThomas2006}, for example.

However, $I(Z; X)$ cannot be calculated analytically.
Thus, we approximate $I(Z; X)$ using the data $\{ x_n \}_{n = 1}^N$ as follows:
\[
  I(Z, X) = \int \mathrm{d}x f(x) \sum_{k = 1}^K
  \frac{\pi_k g_k(x)}{f(x)}
  \log \left( \frac{g_k(x)}{f(x)} \right)
  \approx \frac{1}{N} \sum_{n = 1}^N \sum_{k = 1}^K
  \frac{\pi_k g_k(x_n)}{f(x_n)}
  \log \left( \frac{g_k(x_n)}{f(x_n)} \right) \eqspace ,
\]
where we assume that $0 < f(x_n) < \infty$ holds for all $x_n$.
We call this the MC of the mixture model $f$.

\begin{definition}
  We define the MC of the mixture model $f = \sum_{k = 1}^K \pi_k g_k$
  and that with data weights $\{ w_n \}_{n = 1}^N$
  as the quantities calculated as
  \begin{align}
    \mathrm{MC} (\{ \pi_k, g_k \}_{k = 1}^K; \{ x_n \}_{n = 1}^N)
    \coloneqq \frac{1}{N} \sum_{n = 1}^N \sum_{k = 1}^K
    \frac{\pi_k g_k(x_n)}{f(x_n)}
    \log \left( \frac{g_k(x_n)}{f(x_n)} \right) \label{Eq: definition}
  \end{align}
  and
  \[
    \mathrm{MC} (\{ \pi_k, g_k \}_{k = 1}^K; \{ x_n, w_n \}_{n = 1}^N)
    \coloneqq \frac{1}{\sum_{n' = 1}^N w_{n'}} \sum_{n = 1}^N w_n \sum_{k = 1}^K
    \frac{\pi_k g_k(x_n)}{f(x_n)}
    \log \left( \frac{g_k(x_n)}{f(x_n)} \right) \eqspace ,
  \]
  respectively.
\end{definition}
The weighted version of MC is used in Section \ref{Sec: decomposition}.

Note that there are other ways to approximate $I(Z; X)$;
we adopt the form of Equation (\ref{Eq: definition})
because it has the decomposition property shown in Section \ref{Sec: decomposition}.
See also methods to approximate the entropy of the mixture model
\citep{Huber+2008, KolchinskyTracey2017}
that can also be applied to approximate $I(Z; X)$.

In practice, only the data $\{ x_n \}_{n = 1}^N$
can be obtained without the underlying distribution $f$.
Then, we estimate $\hat{K}$ and $\{\hat{\pi}_k, \hat{g}_k \}_{k = 1}^{\hat{K}}$
from the data $\{ x_n \}_{n = 1}^N$ and further approximate the MC as
\[
  \mathrm{MC} (\{ \pi_k, g_k \}_{k = 1}^K; \{ x_n \}_{n = 1}^N)
  \approx
  \mathrm{MC} (\{ \hat{\pi}_k, \hat{g}_k \}_{k = 1}^{\hat{K}}; \{ x_n \}_{n = 1}^N) \eqspace .
\]

\subsection{Examples}

In this subsection, we discuss some examples of MC to understand its notions.

\subsubsection{MC with different overlaps}

First, we set $N = 600$
and generated the data $x_1, \dots, x_{600} \in \mathbb{R}^2$ as follows.
\begin{align*}
  x_n \sim
  \begin{cases}
    \mathcal{N}(x_n | \mu = [0, 0]^\top, \Sigma = I_2)      & (1 \leq n \leq 300) \eqspace ,   \\
    \mathcal{N}(x_n | \mu = [\alpha, 0]^\top, \Sigma = I_2) & (301 \leq n \leq 600) \eqspace ,
  \end{cases}
\end{align*}
where $\mathcal{N}(x | \mu, \Sigma)$ denotes
a multivariate normal distribution with mean $\mu$ and covariance $\Sigma$,
$I_d$ denotes a $d$-dimensional identity matrix,
and $\alpha \in \mathbb{R}$ is the parameter
that determines the degree of overlap between two components.

By varying the value of $\alpha$ among 0, 0.6, \dots, 6.0,
we generated the data and measured the MC
by setting $\pi_1, \pi_2 = 1 / 2$ and $g_1, g_2$ as the actual distributions.
The exponential of the MC for each $\alpha$ is plotted in Figure \ref{Fig: examples}(a).
It is evident from the figure that the MC smoothly increases from 1.0 to 2.0
as the two components become isolated.

\subsubsection{MC with different mixture biases}

Next, we set $N = 600$
and generated the data $x_1, \dots, x_{600} \in \mathbb{R}^2$ as follows:
\begin{align*}
  x_n \sim
  \begin{cases}
    \mathcal{N}(x | \mu = [0, 0]^\top, \Sigma = I_2) & (1 \leq n \leq 300 + \alpha) \eqspace ,   \\
    \mathcal{N}(x | \mu = [6, 0]^\top, \Sigma = I_2) & (301 + \alpha \leq n \leq 600) \eqspace ,
  \end{cases}
\end{align*}
where $\alpha \in \{0, \dots, 300 \}$ is the parameter
that determines the degree of bias between the proportion of two components.

\begin{figure}[htbp]
  \centering
  \subfigure[MC with different overlaps]{
    \includegraphics[width=0.45\columnwidth]{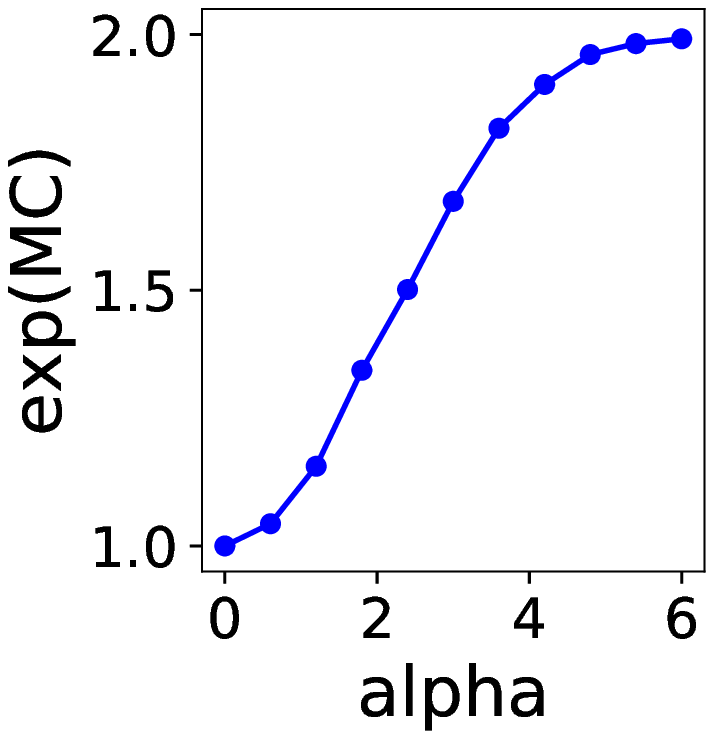}
  }
  \subfigure[MC with different mixture biases]{
    \includegraphics[width=0.45\columnwidth]{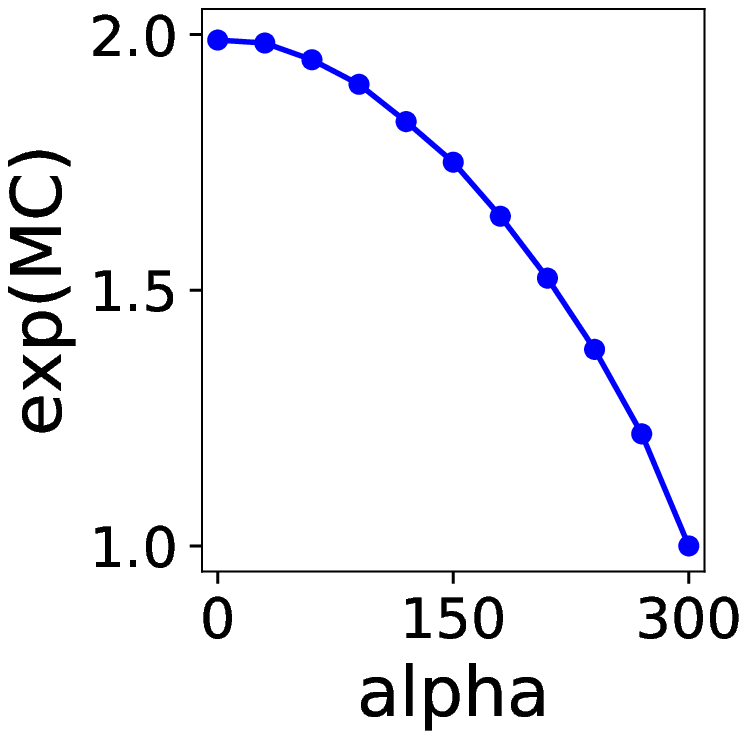}
  }
  \caption{
    Relation between the parameter $\alpha$ and the exponential of the MC.
    \label{Fig: examples}
  }
\end{figure}

By varying $\alpha$ among 0, 30, \dots, 300,
we generated the data and measured the MC
by setting $\pi_1 = (300 + \alpha) / 600, \pi_2 = (300 - \alpha) / 600$
and $g_1, g_2$ as the actual distributions.
The exponential of the MC for each $\alpha$ is plotted in Figure \ref{Fig: examples}(b).
It is evident from the figure that the MC smoothly decreases from 2.0 to 1.0
as the balance becomes biased.

\subsection{Theoretical properties \label{Subsec: MC theoretical}}

In this subsection, we discuss the theoretical properties of MC.
For simplicity, we does not consider the data weights here.
Their proofs are described in Appendix \ref{Sec: proofs}.

First, we show that if the components entirely overlap, MC becomes 0.
\begin{proposition} \label{Prop: entirely overlap}
  If $g_1 = \dots = g_K = g$,
  then
  \[
    \mathrm{MC}(\{ \pi_k, g_k \}_{k = 1}^K; \{ x_n \}_{n = 1}^N) = 0 \eqspace .
  \]
\end{proposition}

Next, we show the relation between MC and mixture size $K$.
If the components are entirely separated and balanced, MC becomes $\log K$.
\begin{proposition} \label{Prop: entirely separated}
  If there is only one index $k$ that satisfies $g_k(x_n) > 0$
  for all $x_n$
  and $\#\{ x_n \mid g_k(x_n) > 0 \} / N = \pi_k $ for all $k$, then
  \[
    \mathrm{MC}(\{ \pi_k, g_k \}_{k = 1}^K; \{ x_n \}_{n = 1}^N) = H (Z) \eqspace .
  \]
  In particular, if $\pi_1 = \dots = \pi_K = 1 / K$, then
  \[
    \mathrm{MC}(\{ \pi_k , g_k \}_{k = 1}^K; \{ x_n \}_{n = 1}^N) = \log K \eqspace .
  \]
\end{proposition}

We also show that if the proportions $\{ \pi_k \}_{k = 1}^K$ are estimated
by maximizing the logarithm of the likelihood, 0 and $\log K$
are the lower and upper bounds of MC, respectively.
\begin{proposition} \label{Prop: bounds}
  If $\{ \pi_k \}_{k = 1}^K$ are an optimal solution of the following problem:
  \begin{align*}
    \max \quad \sum_{n = 1}^N \log \left( \sum_{k = 1}^K \pi'_k g_k(x_n) \right) \quad
    \mathrm{s.t.} \quad \pi'_1, \dots, \pi'_K \geq 0, \ \sum_{k = 1}^K \pi'_k = 1 \eqspace ,
  \end{align*}
  then MC satisfies
  \[
    0 \leq \mathrm{MC}(\{ \pi_k, g_k \}_{k = 1}^K; \{ x_n \}_{n = 1}^N) \leq \log K \eqspace .
  \]
\end{proposition}

Finally, we show that the value of MC is invariant
with the representation of the mixture distribution.
For example, consider the following three mixture distributions:
\begin{align*}
  f_1 (x) & = \frac{1}{2} g_1(x) + \frac{1}{2} g_2(x) \eqspace ,                                        \\
  f_2 (x) & = \frac{1}{2} g_1(x) + \frac{1}{4} g_2(x) + \frac{1}{4} g_2(x) \eqspace ,                   \\
  f_3 (x) & = \frac{1}{2} g_1(x) + \frac{1}{4} g_2(x) + \frac{1}{4} g_2(x) + 0 \cdot g_3 (x) \eqspace .
\end{align*}
In $f_2$ and $f_3$, we need to manually remove the redundant components
and regard the mixture size as two \citep{McLachlanPeel2000}.
On the other hand, the following property indicates that the MCs for
$f_1, f_2$, and $f_3$ are the same;
thus, we need not to care about their differences in evaluating MC.
\begin{proposition} \label{Prop: invariance}
  If two mixture distributions $\sum_{k = 1}^K \pi_k g_k$ and $\sum_{k' = 1}^{K'} \pi'_{k'} g'_{k'}$
  are equivalent in the sense that
  \[
    \sum_{k = 1}^K \pi_k \delta (\cdot, g_k)
    = \sum_{k' = 1}^{K'} \pi'_{k'} \delta (\cdot, g'_{k'}) \eqspace ,
  \]
  where $\delta (\cdot, g)$ is the Kronecker's delta function
  on a function space containing $g_1, \dots, g_K$ and $g'_1, \dots, g'_{K'}$, then
  \[
    \mathrm{MC}(\{ \pi_k, g_k \}_{k = 1}^K; \{ x_n \}_{n = 1}^N)
    = \mathrm{MC}(\{ \pi'_{k'}, g'_{k'} \}_{k' = 1}^{K'}; \{ x_n \}_{n = 1}^N) \eqspace .
  \]
\end{proposition}

\section{Decomposition of MC \label{Sec: decomposition}}

In this section, we discuss a method to decompose MC along the hierarchies in mixture models;
this can help us in analyzing the structures in more detail.

\begin{figure}[htbp]
  \centering
  \includegraphics{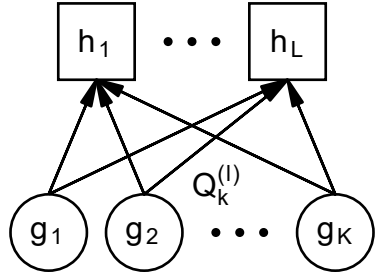}
  \caption{Hierarchy in the mixture model. \label{Fig: hierarchy}}
\end{figure}

Consider that the mixture distribution $f$ has a two-stage hierarchy,
as shown in Figure \ref{Fig: hierarchy}.
It has $K$ components $\{ g_k \}_{k=1}^K$ on the lower side
and $L$ components $\{ h_l \}_{l = 1}^L$ on the upper side,
where $\{ g_k \}_{k = 1}^K$ denote the probability distributions
and $\{ h_l \}_{l = 1}^L$ denote their mixture distributions, respectively.
We construct the hierarchy as follows.
First, we estimate the distribution $f = \sum_{k = 1}^K \pi_k g_k$.
Then, we obtain $\{ h_l \}_{l = 1}^L$ by partitioning (or clustering) the lower components into $L$ groups.
Formally, we denote $Q^{(l)}_k \in \mathbb{R}_{\geq 0}$
as the proportion of the lower component $k \in \{1, \dots, K\}$
that belongs to the upper component $l$,
which satisfies $\sum_{l = 1}^L Q^{(l)}_k = 1$ for all $k$.
Then, we derive $\{ h_l \}_{l = 1}^L$ by rewriting $f = \sum_{k = 1}^K \pi_k g_k$ as
\[
  f(x) = \sum_{k = 1}^K \pi_k g_k(x)
  = \sum_{k = 1}^K \left( \sum_{l = 1}^L Q^{(l)}_k \right) \pi_k g_k(x)
  = \sum_{l=1}^L \rho_l h_l(x) \eqspace ,
\]
where
\[
  \rho_l \coloneqq \sum_{k=1}^K Q^{(l)}_k \pi_k \eqspace , \quad
  h_l (x) \coloneqq \sum_{k=1}^K \phi^{(l)}_k g_k(x) \eqspace , \quad
  \phi^{(l)}_k \coloneqq \frac{Q^{(l)}_k \pi_k}{\rho_l} \eqspace .
\]

According to the hierarchy, we can decompose the MC.
\begin{theorem} \label{Thm: decomposition}
  We can decompose the MC as follows:
  \begin{multline*}
    \mathrm{MC}(\{ \pi_k, g_k \}_{k = 1}^K; \{ x_n \}_{n = 1}^N) \\
    = \mathrm{MC}(\{ \rho_l, h_l \}_{l = 1}^L; \{ x_n \}_{n = 1}^N)
    + \sum_{l=1}^L W_l \cdot
    \mathrm{MC}(\{ \phi^{(l)}_k, g_k \}_{k = 1}^K; \{ x_n, w^{(l)}_n \}_{n = 1}^N) \eqspace ,
  \end{multline*}
  where
  \[
    w^{(l)}_n = \frac{\rho_l h_l(x_n)}{f(x_n)}
    = \frac{\rho_l h_l (x_n)}{\sum_{l' = 1}^L \rho_{l'} h_{l'} (x_n)} \eqspace , \quad
    W_l = \frac{1}{N} \sum_{n = 1}^N w^{(l)}_n \eqspace .
  \]
\end{theorem}
The proof is described in Appendix \ref{Sec: proofs}.

For notational simplicity, we will use the following terms:
\begin{itemize}
  \item $\mathrm{MC}(\{ \pi_k, g_k \}_{k = 1}^K; \{ x_n \}_{n = 1}^N)$:
        MC(total),
  \item $\mathrm{MC}(\{ \rho_l, h_l \}_{l = 1}^L; \{ x_n \}_{n = 1}^N)$:
        MC(interaction),
  \item $W_l \cdot \mathrm{MC}(\{ \phi^{(l)}_k, g_k \}_{k = 1}^K; \{ x_n, w^{(l)}_n \}_{n = 1}^N)$:
        Contribution(component~$l$),
  \item $W_l$:
        W(component~$l$),
  \item $\mathrm{MC}(\{ \phi^{(l)}_k, g_k \}_{k = 1}^K; \{ x_n, w^{(l)}_n \}_{n = 1}^N)$:
        MC(component~$l$).
\end{itemize}
Then, we can rewrite Theorem \ref{Thm: decomposition} as
\begin{gather*}
  \mbox{MC(total)} =
  \mbox{MC(interaction)} + \sum_{l = 1}^L \mbox{Contribution(component~$l$)} \eqspace , \\
  \mbox{Contribution(component~$l$)} =
  \mbox{W(component~$l$)} \cdot \mbox{MC(component~$l$)} \eqspace .
\end{gather*}

In Theorem \ref{Thm: decomposition},
the MC of the entire structure (MC(total)) is decomposed
into a sum of the MC among the upper components (MC(interaction))
and their respective contributions (Contribution(component~$l$)).
Contribution(component~$l$) is further decomposed into a product
of the weight (W(component~$l$)) and complexity (MC(component~$l$)) of the component.
Because $w^{(l)}_n$ denotes the weight of $x_n$ that belongs to component~$l$,
its sum W(component~$l$) represents the total weights of the data contained in it.
Also, MC(component~$l$) denotes the clustering structures in component~$l$
considering the data weights.

\begin{figure}[htbp]
  \centering
  \subfigure[MC(total)]{
    \includegraphics[width=0.45\columnwidth]{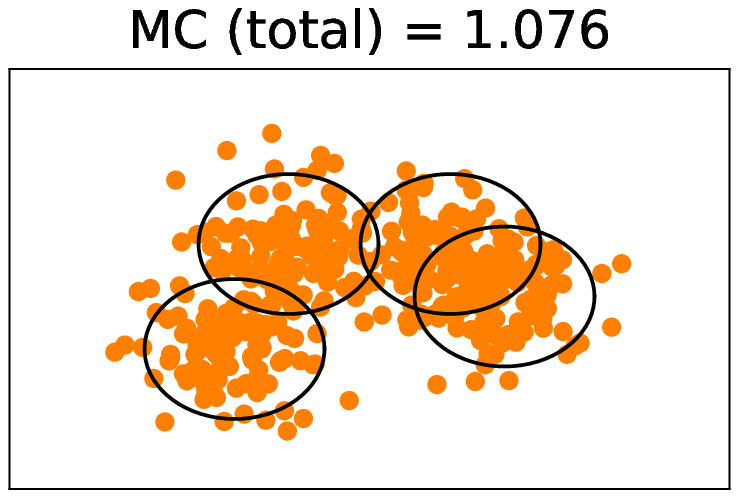}
  }
  \subfigure[MC(interaction)]{
    \includegraphics[width=0.45\columnwidth]{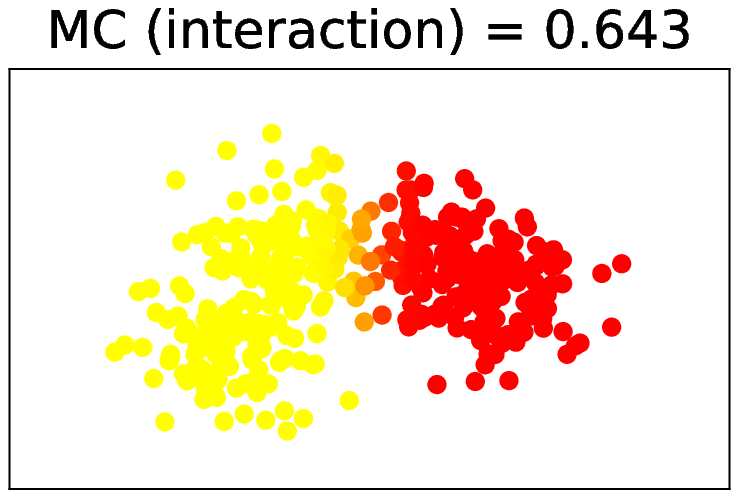}
  }
  \subfigure[MC(component 1)]{
    \includegraphics[width=0.45\columnwidth]{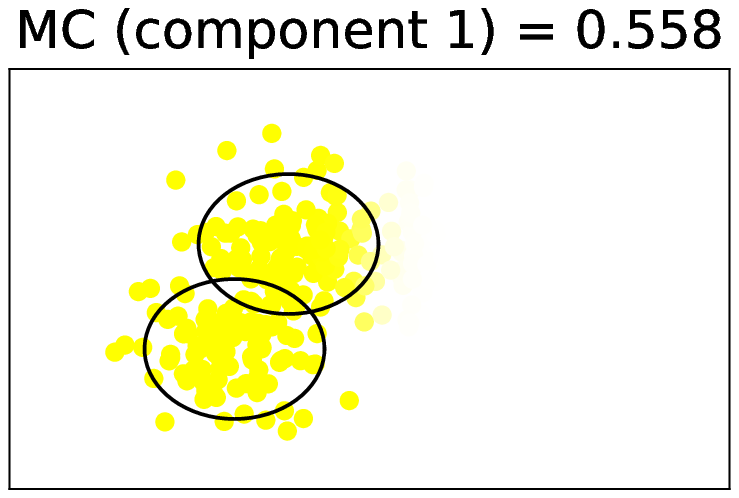}
  }
  \subfigure[MC(component 2)]{
    \includegraphics[width=0.45\columnwidth]{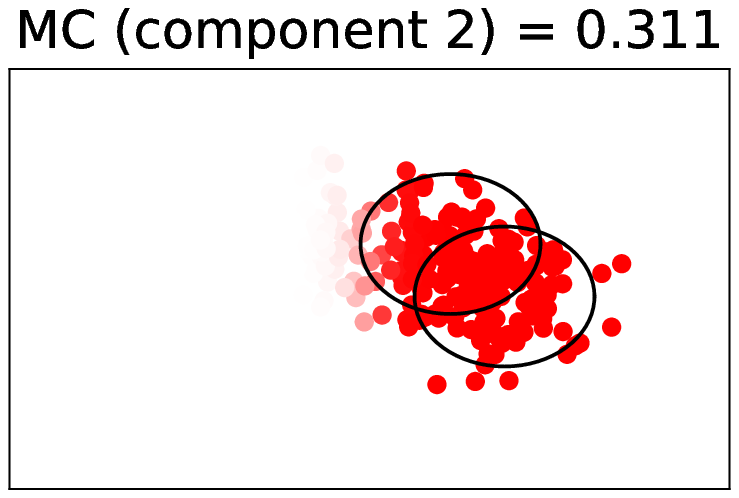}
  }
  \caption{
    Example of the decomposition of MC.
    The data's color in (b) and thickness in (c) and (d) correspond to the data weights $w^{(l)}_n$.
    \label{Fig: decomposition}
  }
\end{figure}

An example of  the decomposition is illustrated in Figure \ref{Fig: decomposition}
and Table \ref{Tab: decomposition}.
In this example, there are $K = 4$ lower components generated from a Gaussian mixture model;
additionally, there are $L = 2$ upper components on the left and right sides.
By decomposing MC(total),
we can evaluate the complexities in the local structures as well as those in the entire structure.

\begin{table}[htbp]
  \centering
  \begin{tabular}{r|cc}
    \hline
                                & component 1               & component 2 \\
    \hline
    MC(total)                   & \multicolumn{2}{c}{1.076}               \\ \hline
    MC(interaction)             & \multicolumn{2}{c}{0.643}               \\
    Contribution(component $l$) & 0.277                     & 0.157       \\ \hline
    W(component $l$)            & 0.496                     & 0.504       \\
    MC(component $l$)           & 0.558                     & 0.311       \\
    \hline
  \end{tabular}
  \caption{Quantities in the example of the decomposition. \label{Tab: decomposition}}
\end{table}

\section{Application to clustering change detection \label{Sec: application}}

In this section, we propose methods to apply the MC to clustering change detection problems.
Formally speaking, given the dataset
$\mathcal{X} \coloneqq \{ \{ x_{n, t} \}_{n = 1}^N \mid t \in 1, \dots, T \}$,
where $t$ denotes the time and $\{ x_{n, t} \}_{n = 1}^N$ denote the data generated at each $t$,
we consider the problem of monitoring the changes in the clustering structures
over $t = 1, \dots, T$.

First, we briefly summarize the method
named sequential dynamic model selection (SDMS) \citep{HiraiYamanishi2012}
that has addressed this problem.
Then, we introduce our ideas and discuss the differences between SDMS.

Hereafter, we assume that the data points $x_{n, t}$ are $d$-dimensional vectors
and consider a Gaussian mixture model
\[
  f_t (x) = \sum_{k = 1}^{K_t} \pi_{k, t} \,
  \mathcal{N}( x | \mu_{k, t}, \Sigma_{k, t})
\]
for each $t$.

\subsection{Sequential dynamic model selection \label{Subsec: SDMS}}

SDMS is an algorithm that is used to sequentially estimate models and find changes.
In clustering change detection problems,
it sequentially estimates the mixture sizes $\hat{K}_t$ and parameters
$\theta_{\hat{K}_t} \coloneqq
  \{ \hat{\pi}_{k, t}, \hat{\mu}_{k, t}, \hat{\Sigma}_{k, t} \}_{k = 1}^{\hat{K}_t}$
and finds model changes as changes in $\hat{K}_t$.

The estimation procedures are explained below.
First, depending on the estimated mixture size at the last time point $\hat{K}_{t - 1}$,
we set the candidate for $K_t$.
Then, for each $K_t$ in the candidate, we estimate the parameters
$\theta_{K_t}$ from the data $\{ x_{n, t} \}_{n = 1}^N$ and calculate a cost function
$\mathcal{L}_{\mathrm{SDMS}}(\{ x_{n, t} \}_{n = 1}^N; K_t, \theta_{K_t}, \hat{K}_{t - 1})$.
Finally, we select $K_t$ as the mixture size that minimizes the costs.
The candidate of $K_t$ are set as
\[
  \{1, \dots, K_{\mathrm{max}} \}
\]
at $t = 1$, and
\[
  \{K_{t - 1} - 1, K_{t - 1}, K_{t - 1} + 1\} \cap \{1, \dots, K_{\mathrm{max}} \}
\]
at $t \geq 2$, where $K_{\mathrm{max}}$ is a pre-defined parameter.
The cost function denotes the sum of the code length functions of the model and model changes
given by
\[
  \mathcal{L}_{\mathrm{SDMS}}(\{ x_{n, t} \}_{n = 1}^N; K_t, \theta_{K_t}, \hat{K}_{t - 1})
  = \mathcal{L}_{\mathrm{model}}(\{ x_{n, t} \}_{n = 1}^N; K_t, \theta_{K_t})
  + \mathcal{L}_{\mathrm{change}}(K_t | \hat{K}_{t - 1}) \eqspace .
\]

\paragraph{Code length of the model}

The score $\mathcal{L}_{\mathrm{model}}(\{ x_n \}_{n = 1}^N; K, \theta_{K})$
denotes a sum of the logarithm of the likelihood functions and penalty terms
corresponding to the complexity of the model.
In this study, we consider two likelihood functions and four penalty terms.
For the (logarithm of) likelihood functions,
we consider the \emph{observed likelihood} $L(\{ x_n \}_{n = 1}^N; \theta_K)$
and \emph{complete likelihood} $L(\{ x_n, z_n \}_{n = 1}^N; \theta_K)$, given by
\begin{gather*}
  L(\{ x_n \}_{n = 1}^N; \theta_K)
  \coloneqq \sum_{n = 1}^N \log p (X = x_n)
  = \sum_{n = 1}^N \left( \sum_{k = 1}^K \pi_k \ \mathcal{N} (x_n | \mu_k, \Sigma_k) \right) \eqspace ,  \\
  L(\{ x_n, z_n \}_{n = 1}^N; \theta_K)
  \coloneqq \sum_{n = 1}^N \log p (X = x_n, Z = z_n)
  = \sum_{n = 1}^N \log \left( \pi_{z_n} \ \mathcal{N} (x_n | \mu_{z_n}, \Sigma_{z_n}) \right) \eqspace ,
\end{gather*}
where $\{ z_n \}_{n = 1}^N$ are the latent variables for the data estimated by
\[
  z_n \coloneqq \underset{z \in 1, \dots, K}{\mathrm{argmax}} \ p(Z = z | X = x_n) \eqspace .
\]
They correspond to the likelihood of the observed data and complete data, respectively;
the former is used to determine the mixture size,
and the latter is used to determine the cluster size
under the assumption that it is equal to the mixture size.
For the penalty terms, we consider AIC \citep{Akaike1974}, BIC \citep{Schwarz1978},
NML \citep{HiraiYamanishi2013, HiraiYamanishi2019}, and DNML \citep{Wu+2017, Yamanishi+2019}.
By combining the log-likelihood and the penalty terms,
we consider the following six scores:
\begin{itemize}
  \item AIC with observed likelihood (AIC):
        $- L(\{ x_n \}_{n = 1}^N; \theta_K) + D$,
  \item AIC with complete likelihood (AIC+comp):
        $- L(\{ x_n, z_n \}_{n = 1}^N; \theta_K) + D$,
  \item BIC with observed likelihood (BIC):
        $- L(\{ x_n \}_{n = 1}^N; \theta_K) + (D / 2) \log N$,
  \item BIC with complete likelihood (BIC+comp):
        $- L(\{ x_n, z_n \}_{n = 1}^N; \theta_K) + (D / 2) \log N$,
  \item NML:
        $- L(\{ x_n, z_n \}_{n = 1}^N; \theta_K) + \log C_\mathrm{NML}(N, K)$,
  \item DNML:
        $- L(\{ x_n, z_n \}_{n = 1}^N; \theta_K) + \log C_\mathrm{DNML}(N, \{ z_n \}_{n = 1}^N, K)$,
\end{itemize}
where $D \coloneqq (K - 1) + d (d + 3) / 2$ denotes the number of the free parameters
required to represent a Gaussian mixture model;
$C_\mathrm{NML}(N, K)$ and $C_\mathrm{DNML}(N, \{ z_n \}_{n = 1}^N, K)$
denote the parametric complexities.
In our experiments, we estimated the parameter $\theta_{K}$
by conducting the EM algorithm \citep{Dempster+1977} implemented
in the Scikit-learn package \citep{Pedregosa+2011} ten times
and choose the best parameter that minimized each score.
Note that in NML and DNML, we only considered the complete likelihood functions
because only the methods to calculate their parametric complexities are known.

\paragraph{Code length for model change}

The code length for the model change
$\mathcal{L}_{\mathrm{change}}(K_t | \hat{K}_{t - 1})$
can be written as
\[
  \mathcal{L}_{\mathrm{change}}(K_t | \hat{K}_{t - 1}) = - \log p(K_t | \hat{K}_{t - 1}, \beta) \eqspace ,
\]
where $p(K_t | \hat{K}_{t - 1}, \beta)$ denotes the probability of the model change, defined as
\[
  p(K_1) \coloneqq 1 / K_\mathrm{max}
\]
for all $K_1$ at $t = 1$ and
\[
  p (K_t | \hat{K}_{t - 1}, \beta)
  = \begin{cases}
    1 - \beta / 2 &
    \mbox{if } K_t = \hat{K}_{t - 1} \mbox{ and }
    \hat{K}_{t - 1} \in \{ 1, K_\mathrm{max} \} \eqspace ,    \\
    1 - \beta     &
    \mbox{if } K_t = \hat{K}_{t - 1} \mbox{ and }
    \hat{K}_{t - 1} \notin \{ 1, K_\mathrm{max} \} \eqspace , \\
    \beta / 2     &
    \mbox{if } K_t \neq \hat{K}_{t - 1}
  \end{cases}
\]
at $t \geq 2$; additionally, $0 < \beta < 1$ is a predefined parameter.
In our experiments, we fixed $\beta = 0.01$.

\subsection{Tracking MC \label{Subsec: Track MC}}

In SDMS, clustering changes are detected as the changes of the mixture size or cluster size $K$;
because it is discrete, the changes have been considered to be abrupt.
Then, we propose to track MC instead of $K$ while estimating the parameters by SDMS.
Because MC takes a real value,
monitoring it is more suitable for observing gradual changes than monitoring $K$.
The algorithm for tracking MC is explained in Algorithm \ref{Alg: Tracking MC}.

\begin{algorithm}[ht]
  \caption{Tracking MC \label{Alg: Tracking MC}}
  \begin{algorithmic}[1]
    \REQUIRE A dataset $\mathcal{X} = \{ \{ x_{n, t} \}_{n = 1}^N \mid t \in 1, \dots, T \}$.
    \FOR{$t = 1$ \TO $t = T$}
    \STATE Estimate $\hat{K}_t$ and $\{ \hat{g}_{k, t} \}_{k = 1}^{\hat{K}_t}$
    from the data $\{ x_{n, t} \}_{n = 1}^N$ using SDMS.
    \STATE Calculate
    $
      \mathrm{MC}_t \coloneqq
      \mathrm{MC}(\{ \hat{\pi}_{k, t}, \hat{g}_{k, t} \}_{k = 1}^{\hat{K}_t};
      \{ x_{n, t} \}_{n = 1}^N)
    $.
    \ENDFOR
    \RETURN $\{ \mathrm{MC}_t \}_{t = 1}^T$.
  \end{algorithmic}
\end{algorithm}

\subsection{Tracking MC with its decomposition}

In addition to monitoring the MC of the entire structure,
we also propose an algorithm to track its decomposition.
To accomplish this, we must estimate the upper $L$ components
and their corresponding partitions $Q^{(l)}_{k, t}$ for each $t$.

Here, we assume that the upper $L$ components are common at every $t$
and estimate the partition $Q^{(l)}_{k, t}$ after estimating the lower components at each time.
Specifically, we consider $\mu_{k, t}$ as a point with weights $\pi_{k, t}$
for each $k$ and $t$ and cluster them.
As the clustering algorithm, we modified the fuzzy c-means \citep{Bezdec+1984}
to handle the weighted points.
Formally, we estimated the centers of the upper $L$ components $\tilde{\mu}_l$
and their corresponding partitions $Q^{(l)}_{k, t}$ by minimizing the loss function
\begin{align*}
  \sum_{t, k} \pi_{k, t} \sum_{l = 1}^L \left( Q^{(l)}_{k, t} \right)^m
  \| \mu_{k, t} - \tilde{\mu}_l \|^2,
\end{align*}
where $m > 0$ is parameter that determines the fuzziness of the partition.

We estimated $\tilde{\mu}_l$ and $Q^{(l)}_{k, t}$ by minimizing one iteratively while fixing another.
We can formulate the iteration as follows:
\[
  \tilde{\mu}_l =
  \frac{\sum_{k, t} \pi_{k, t} \left( Q^{(l)}_{k, t} \right)^m \mu_{k, t}}
  {\sum_{k, t} \pi_{k, t} \left( Q^{(l)}_{k, t} \right)^m} \eqspace , \quad
  Q^{(l)}_{k, t} =
  \frac{\| \mu_{k, t} - \tilde{\mu}_l \|^{2 / (m - 1)}}
  {\sum_{l' = 1}^L \| \mu_{k, t} - \tilde{\mu}_{l'} \|^{2 / (m - 1)}} \eqspace .
\]

Finally, we present an algorithm to track the MC and its decomposition
in Algorithm \ref{Alg: tracking decomposition}.
We can analyze the structural changes in more detail by evaluating the decomposed values.

\begin{algorithm}[ht]
  \caption{Tracking MC with its decomposition \label{Alg: tracking decomposition}}
  \begin{algorithmic}[1]
    \REQUIRE A dataset $\mathcal{X} = \{ \{ x_{n, t} \}_{n = 1}^N \mid t \in 1, \dots, T \}$,
    parameters $m$ and $L$.
    \STATE
    \STATE \textit{\# Step 1: Estimate lower components.}
    \FOR{$t = 1$ \TO $t = T$}
    \STATE Estimate $\hat{K}_t$ and $\{ \hat{g}_{k, t} \}_{k = 1}^{\hat{K}_t}$
    from the data $\{ x_{n, t} \}_{n = 1}^N$ using SDMS.
    \STATE Calculate
    $
      \mbox{MC(total)}_t \coloneqq
      \mathrm{MC}(\{ \hat{\pi}_{k, t}, \hat{g}_{k, t} \}_{k = 1}^{\hat{K}_t};
      \{ x_{n, t} \}_{n = 1}^N)
    $.
    \ENDFOR
    \STATE
    \STATE \textit{\# Step 2: Estimate upper components and partition.}
    \STATE Estimate the centers $\tilde{\mu}_l$ and the partition $Q^{(l)}_{k, t}$
    using fuzzy c-means.
    \STATE
    \STATE \textit{\# Step 3: Calculate the decomposition of MC.}
    \FOR{$t = 1$ \TO $t = T$}
    \STATE Calculate $\mbox{MC(interaction)}_t$
    defined in Section \ref{Sec: decomposition}.
    \FOR{$l = 1$ \TO $l = L$}
    \STATE Calculate $\mbox{W(component~$l$)}_t$
    defined in Section \ref{Sec: decomposition}.
    \STATE Calculate $\mbox{MC(component~$l$)}_t$
    defined in Section \ref{Sec: decomposition}.
    \ENDFOR
    \ENDFOR
    \RETURN
    $
      \{
      \mbox{MC(total)}_t, \mathrm{MC(interaction)}_t,
      \{\mbox{W(component~$l$)}_t, \mbox{MC(component~$l$)}_t \}_{l = 1}^L
      \}_{t = 1}^T.
    $
  \end{algorithmic}
\end{algorithm}

\section{Experimental results \label{Sec: results}}

In this section, we present the experimental results
that demonstrate the MC's abilities to monitor the clustering changes.
We compare our methods to the monitoring of $K$.

\subsection{Analysis of artificial data}

To reveal the behaviors of MC, we conducted experiments with two artificial datasets
called {\it move Gaussian dataset} and {\it imbalance Gaussian dataset}.
Their experimental designs are discussed below.
First, we generated artificial datasets
$\mathcal{X} = \{ \{ x_{n, t} \}_{n = 1}^N \mid t \in 1, \dots, T \}$
by setting $T = 150$ and $N = 1000$.
The datasets have one transaction period $t = 51, \dots, 100$
in which the data change their clustering structures gradually.
Then, we estimated the MC and $K$ using the methods in Subsections
\ref{Subsec: Track MC} and \ref{Subsec: SDMS} by setting $K_{\mathrm{max}} = 10$.
To compare them, we first created a simple algorithm to detect the changes
from the sequence of MC or $K$.
Then, we compared the abilities of this algorithm
in terms of the speed and accuracy of detecting the change points.
Moreover, to evaluate the abilities to find the changes in the opposite direction,
we performed experiments with the same datasets in the reverse order.

Given a sequence of the MC or $K$ written as $y_1, \dots, y_{150}$,
we constructed an algorithm to detect the change points as follows.
For $t = 10, \dots, 150$, we raised a change alert if
\[
  |\mathrm{median} (y_{t - 9}, \dots, y_{t - 5}) - \mathrm{median} (y_{t - 4}, \dots, y_t)| > 0.01
\]
in the case of MC, and
\[
  \mathrm{median} (y_{t - 9}, \dots, y_{t - 5}) \neq \mathrm{median} (y_{t - 4}, \dots, y_t)
\]
in the case of $K$.
We calculated the medians instead of the means of the subsequences for robustness.
However, to avoid redundant alerts, we neglected them
when the difference between $t$ and the latest alert was less than 5
even if the conditions were satisfied.

To evaluate the quality of the algorithm,
we calculated \emph{Delay} and \emph{False alarm rate} (FAR), defined as
\[
  \mathrm{Delay} \coloneqq \min \left( t^* - 51, 50 \right) \eqspace , \quad
  \mathrm{FAR} \coloneqq
  \frac{\# \{t \in [10, 150] \mid t \notin \mathrm{ACCEPT} \land t \in \mathrm{ALERT}\}}
  {\# \{t \in [10, 150] \mid t \notin \mathrm{ACCEPT} \}} \eqspace ,
\]
where $t^*$ denotes the first time point in the transaction period
when the algorithm generated an alert,
ACCEPT denotes the set of time points when alerts are allowed defined as
$\{ t \mid \exists t - 9, \dots, t \in [51, \dots, 100] \} = [51, 109]$,
and ALERT denotes the set of time points when the algorithm generated alerts.

\subsubsection{Move Gaussian dataset}

\begin{figure}[htbp]
  \centering
  \subfigure[$t = 1$]{
    \includegraphics[width=0.3\columnwidth]{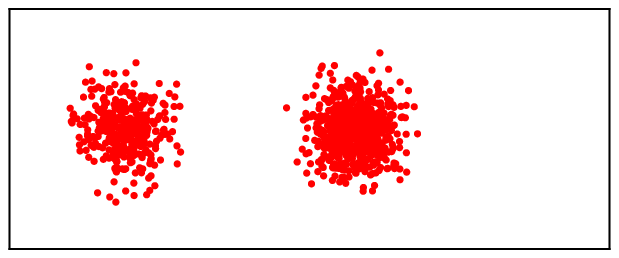}
  }
  \subfigure[$t = 75$]{
    \includegraphics[width=0.3\columnwidth]{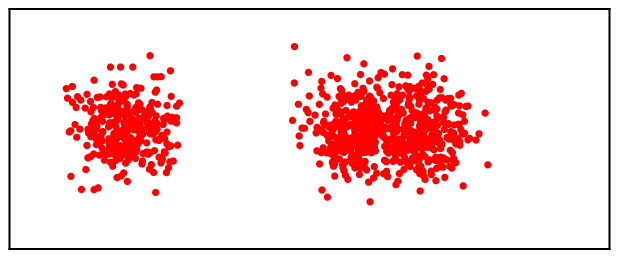}
  }
  \subfigure[$t = 101$]{
    \includegraphics[width=0.3\columnwidth]{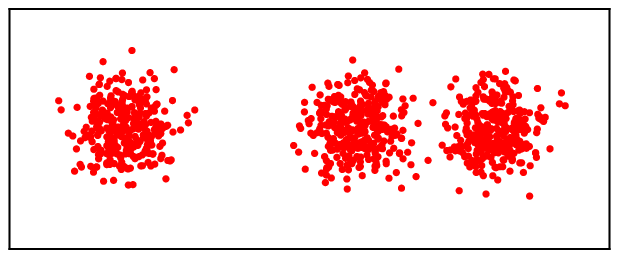}
  }
  \caption{
    Scatter plots of the first and second dimensions of the data
    at $t = 1, 75, 101$ in the move Gaussian dataset.
    \label{Fig: example Move Gaussian}
  }
\end{figure}

The move Gaussian dataset is a set of three-dimensional Gaussian distributions,
whose means move gradually in the transaction period.
Formally, for each $t$, we generated the data $\{ x_{n, t} \}_{n = 1}^{1000}$ as follows:
\[
  x_{n, t} \sim
  \begin{cases}
    \mathcal{N}(x | \mu = [0, 0, 0]^\top, \Sigma = I_3)              & (1 \leq n \leq 333) \eqspace ,    \\
    \mathcal{N}(x | \mu = [10, 0, 0]^\top, \Sigma = I_3)             & (334 \leq n \leq 666) \eqspace ,  \\
    \mathcal{N}(x | \mu = [10 + \alpha(t), 0, 0]^\top, \Sigma = I_3) & (667 \leq n \leq 1000) \eqspace , \\
  \end{cases}
\]
where
\[
  \alpha (t) =
  \begin{cases}
    0             & (1 \leq t \leq 50) \eqspace ,    \\
    0.12 (t - 50) & (51 \leq t \leq 100) \eqspace ,  \\
    6             & (101 \leq t \leq 150) \eqspace . \\
  \end{cases}
\]
The first and second dimensions of some data are visualized
in Figure \ref{Fig: example Move Gaussian}.
In the direction $t = 1 \to 150$, the number of clusters increases from two to three
as the two clusters leave;
in the direction $t = 150 \to 1$, it decreases from three to two
as the two clusters merge.

\begin{figure}[htbp]
  \centering
  \subfigure[$t = 1 \to 150$]{
    \includegraphics[width=0.9\columnwidth]{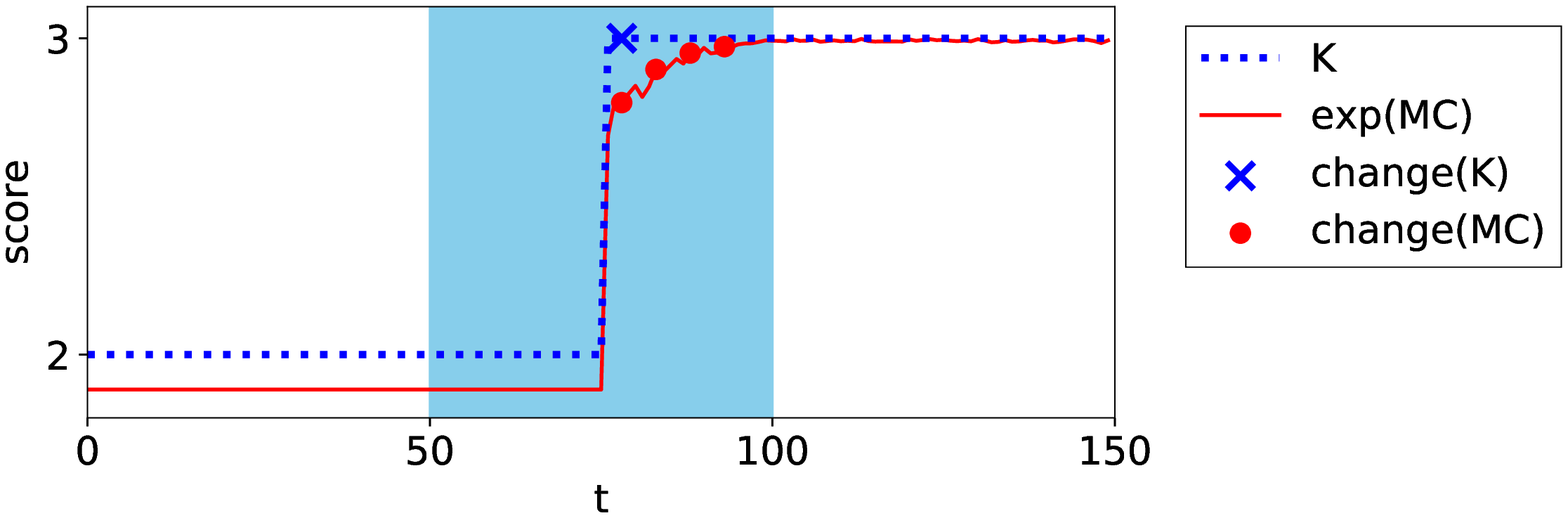}
  }
  \subfigure[$t = 150 \to 1$]{
    \includegraphics[width=0.9\columnwidth]{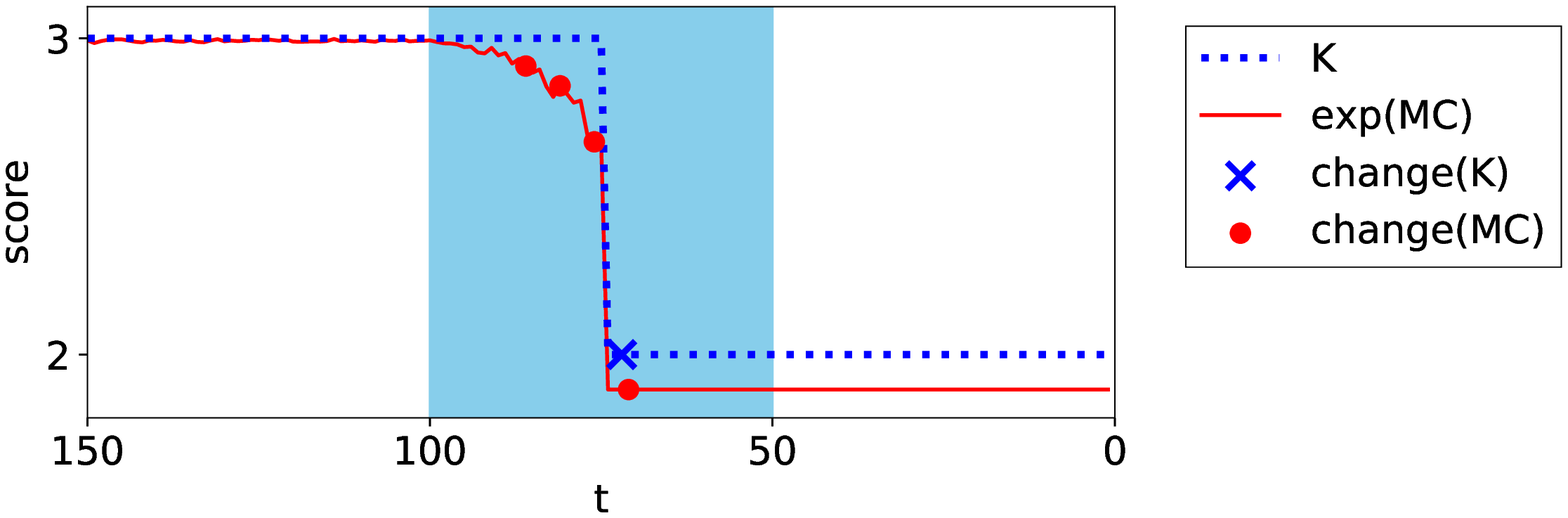}
  }
  \caption{
    Plots of the exponential of MC and $K$ for the move Gaussian dataset.
    The filled area represents the transaction period.
    The markers on the plot represent the alerts in each method.
    \label{Fig: plot Move Gaussian}
  }
\end{figure}

The experiments were performed ten times by randomly generating the datasets;
accordingly, the average performance scores were calculated.
The differences in the scores between the MC and $K$ for each criterion are presented
in Table \ref{Tab: score Move Gaussian};
the estimated MC and $K$ in one trial are proposed in Figure \ref{Fig: plot Move Gaussian}.
This figure illustrates the result of BIC as an example.

With respect to the speed to find changes, in every criterion,
MC performed as well as $K$ in the direction $t = 1 \to 150$;
however, it performed significantly better than $K$ in the direction $t = 150 \to 1$.
The reason for the differing performances is discussed below.
In the direction $t = 1 \to 150$,
the model selection algorithms underestimated the number of components
at the beginning of the transaction period.
In such time points, they ignored the overlapping of the two components
and considered them as one cluster.
Thus, MC, based on such model selection methods, was unable to find the changes earlier than $K$.
However, in the direction $t = 150 \to 1$,
the overlap between the components was correctly estimated at some time points before $K$ changed.
In this case, MC changed smoothly according to the overlap and found changes earlier than $K$.

With respect to the accuracy of finding changes,
MC performed as well as $K$ in terms of FAR.
Additionally, it is evident from Figure \ref{Fig: plot Move Gaussian}
that MC stably estimated the clustering structures.

\begin{table}[htbp]
  \centering
  \begin{tabular}{r|rr|rr}
    \hline
              & \multicolumn{4}{c}{(score of MC) - (score of $K$)}                         \\ \cline{2-5}
              & \multicolumn{2}{c|}{$t = 1 \to 150$}
              & \multicolumn{2}{c}{$t = 150 \to 1$}                                        \\ \cline{2-5}
    criterion & Delay                                              & FAR   & Delay & Far   \\ \hline
    AIC       & 0.0                                                & 0.000 & -20.6 & 0.000 \\
    AIC+comp  & 0.0                                                & 0.000 & -10.9 & 0.000 \\
    BIC       & 0.0                                                & 0.000 & -17.5 & 0.000 \\
    BIC+comp  & 0.0                                                & 0.000 & -8.9  & 0.000 \\
    NML       & 0.0                                                & 0.000 & -7.9  & 0.000 \\
    DNML      & 0.0                                                & 0.000 & -7.7  & 0.000 \\ \hline
  \end{tabular}
  \caption{
    Difference in the average performance score between MC and $K$ for the move Gaussian dataset.
    \label{Tab: score Move Gaussian}
  }
\end{table}

\subsubsection{Imbalance Gaussian dataset}

\begin{figure}[htbp]
  \centering
  \subfigure[$t = 1$]{
    \includegraphics[width=0.3\columnwidth]{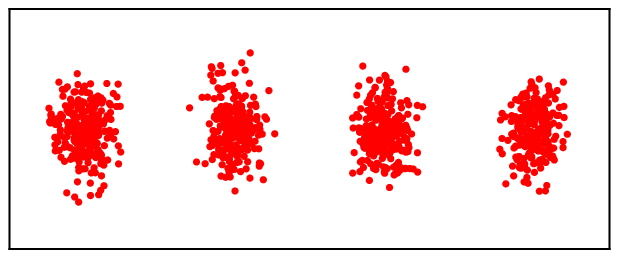}
  }
  \subfigure[$t = 80$]{
    \includegraphics[width=0.3\columnwidth]{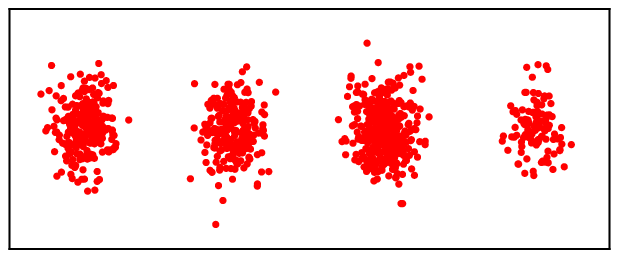}
  }
  \subfigure[$t = 101$]{
    \includegraphics[width=0.3\columnwidth]{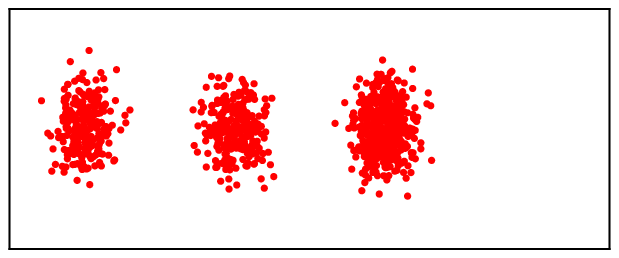}
  }
  \caption{
    Scatter plots of the first and second dimensions of the data
    at $t = 1, 80, 101$ in the imbalance Gaussian dataset.
    \label{Fig: example Imbalance Gaussian}
  }
\end{figure}

The imbalance Gaussian dataset is a set of three-dimensional Gaussian mixture distributions
whose balances change gradually in the transaction period.
Formally, for each $t$, we generated the data $\{ x_{n, t} \}_{n = 1}^{1000}$ as follows:
\[
  x_{n, t} \sim
  \begin{cases}
    \mathcal{N}(x | \mu = [0, 0, 0]^\top, \Sigma = I_3)  & (1 \leq n \leq 250) \eqspace ,                 \\
    \mathcal{N}(x | \mu = [10, 0, 0]^\top, \Sigma = I_3) & (251 \leq n \leq 500) \eqspace ,               \\
    \mathcal{N}(x | \mu = [20, 0, 0]^\top, \Sigma = I_3) & (501 \leq n \leq 750 + \alpha (t)) \eqspace ,  \\
    \mathcal{N}(x | \mu = [30, 0, 0]^\top, \Sigma = I_3) & (751 + \alpha (t) \leq n \leq 1000) \eqspace ,
  \end{cases}
\]
where
\[
  \alpha (t) = \begin{cases}
    0          & (1 \leq t \leq 50) \eqspace ,    \\
    5 (t - 51) & (51 \leq t \leq 100) \eqspace ,  \\
    250        & (101 \leq t \leq 150) \eqspace .
  \end{cases}
\]
The first and second dimensions of some data are visualized
in Figure \ref{Fig: example Imbalance Gaussian}.
In the direction $t = 1 \to 150$, the number of clusters decreases from four to three
as the edge cluster disappears.
In the direction $t =150 \to 1$, it increases from three to four
as the edge cluster emerges.

\begin{figure}[htbp]
  \centering
  \subfigure[$t = 1 \to 150$]{
    \includegraphics[width=0.9\columnwidth]{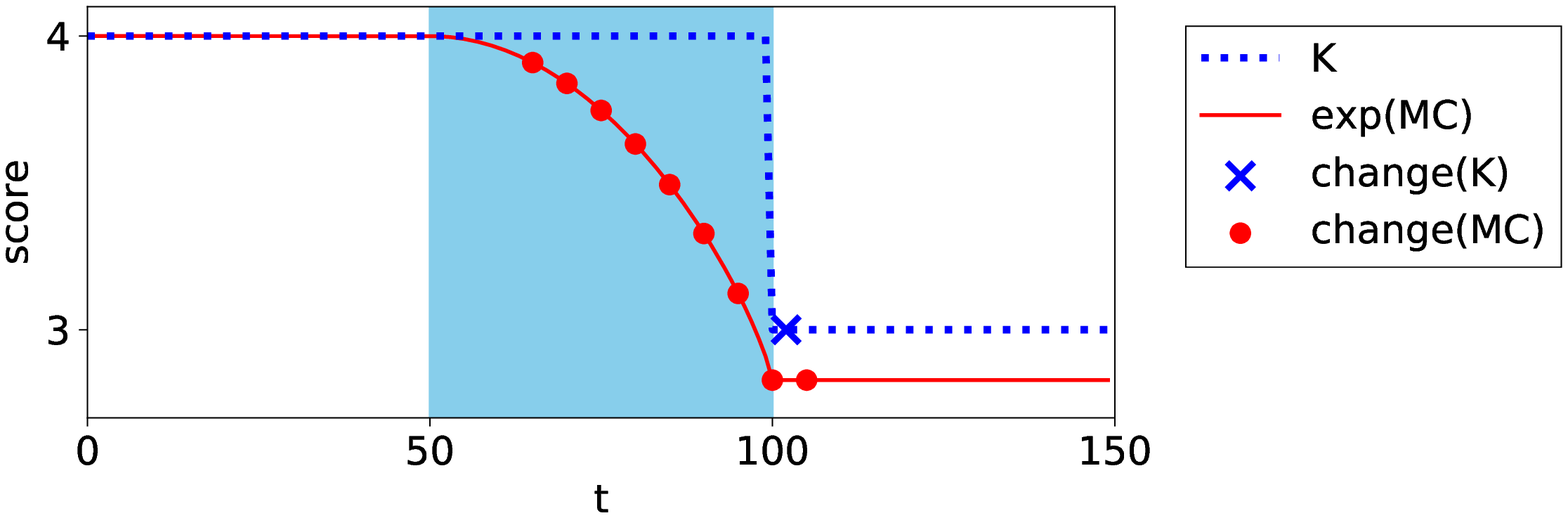}
  }
  \subfigure[$t = 150 \to 1$]{
    \includegraphics[width=0.9\columnwidth]{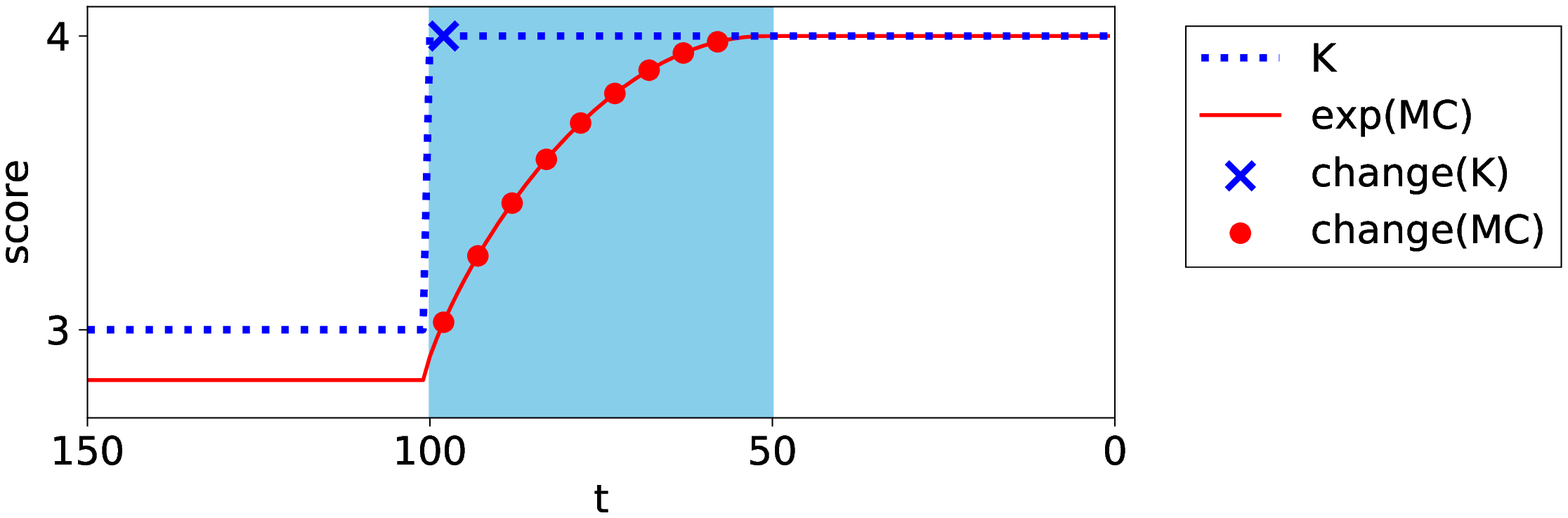}
  }
  \caption{
    Plots of the exponential of MC and $K$ for the imbalance Gaussian dataset.
    The filled area represents the transaction period.
    The markers on the plot represent the alerts in each method.
    \label{Fig: plot Imbalance Gaussian}
  }
\end{figure}

The experiments were performed ten times by randomly generating datasets;
accordingly, the average performance scores were calculated.
The difference in the scores between the MC and $K$ for each criterion are listed
in Table \ref{Tab: score Imbalance Gaussian}.
The estimated MC and $K$ in one trial are plotted in Figure \ref{Fig: plot Imbalance Gaussian}.
This figure illustrates the result of BIC as an example.

In terms of the speed to find changes, in every model selection method,
MC performed significantly better than $K$ in the direction $t = 1 \to 150$;
however, MC performed as well as $K$ in the direction $t = 150 \to 1$.
The reason for the differing performances is discussed below.
In the transaction period,
all model selection methods counted the minor components as independent clusters.
Then, in the direction $t = 1 \to 150$, MC changed smoothly according to the imbalance
and determined the changes earlier than $K$.
In the direction $t = 150 \to 1$, $K$ increased significantly early in the transaction period.
Then, MC increased along with $K$ and determined the changes simultaneously.

In terms of the accuracy of finding changes, MC performed as well as $K$ in terms of FAR.
Additionally, it is evident from Figure \ref{Fig: plot Imbalance Gaussian}
that MC stably estimated the clustering structures.

\begin{table}[htbp]
  \centering
  \begin{tabular}{r|rr|rr}
    \hline
              & \multicolumn{4}{c}{(score of MC) - (score of $K$)}                         \\ \cline{2-5}
              & \multicolumn{2}{c|}{$t = 1 \to 150$}
              & \multicolumn{2}{c}{$t = 150 \to 1$}                                        \\ \cline{2 - 5}
    criterion & Delay                                              & FAR   & Delay & Far   \\ \hline
    AIC       & -30.2                                              & 0.010 & -4.6  & 0.000 \\
    AIC+comp  & -34.0                                              & 0.000 & 0.0   & 0.000 \\
    BIC       & -34.0                                              & 0.000 & 0.0   & 0.000 \\
    BIC+comp  & -34.0                                              & 0.000 & 0.0   & 0.000 \\
    NML       & -34.0                                              & 0.000 & 0.0   & 0.000 \\
    DNML      & -34.0                                              & 0.000 & 0.0   & 0.000 \\ \hline
  \end{tabular}
  \caption{
    Differences in the average performance score between MC and $K$ for the imbalance Gaussian dataset.
    \label{Tab: score Imbalance Gaussian}
  }
\end{table}

\subsection{Analysis of real data}

We analyzed two types of real data named the \emph{beer dataset} and \emph{house dataset},
which are summarized in Table \ref{Tab: dataset}.
In the following subsections, we discuss the detail of the datasets and results of the experiments.

\begin{table}[ht]
  \centering
  \begin{tabular}{c|c|c|c|c} \hline
    dataset & $T$ & $N_t$ & $d$ & description                              \\ \hline
    beer    & 92  & 3185  & 16  & purchase data of beer.                   \\ \hline
    house   & 96  & 4326  & 3   & electricity consumption data in a house. \\ \hline
  \end{tabular}
  \caption{Summary of the dataset \label{Tab: dataset}}
\end{table}

\subsubsection{Beer dataset}

We discuss the results of the beer dataset, obtained from Hakuhodo, Inc. and M-CUBE, Inc.
The dataset comprises the records of customer's beer purchases
from November 1st, 2010 to January 31th, 2011.
The dataset $\mathcal{X}$ is constructed as follows.
The time unit is a day.
For each day $t \in \{\tau, \dots, T \}$,
$x_{n, t} \in \mathbb{R}^d$
denotes the $n$-th customer's consumption of the beer
from time $t - \tau + 1$ to $t$, where we set $\tau = 14$.
The dimension $d$ of the vector is 16, which correspond to the consumptions of the following drink:
\begin{itemize}
  \item beer(A), \dots, beer(F): beer with brand name A, \dots, F.
  \item beer(other): beer with other brands.
  \item beerlike(A), \dots, beerlike(H): beer-like drink with brand name A, \dots, H.
  \item beerlike(other): beer-like drink with other brands.
\end{itemize}

\begin{figure}[htbp]
  \centering
  \includegraphics[width=0.9\columnwidth]{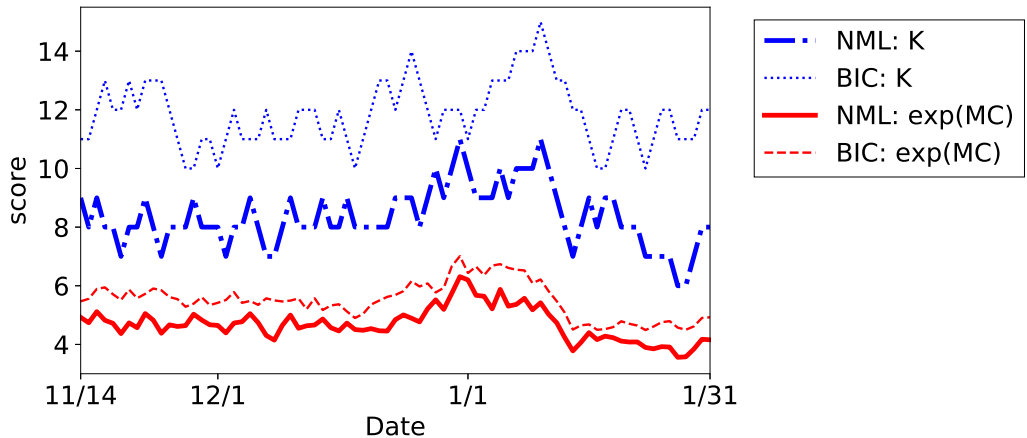}
  \caption{
    Plots of the sequences of the MC and $K$ in the beer dataset.
    \label{Fig: MC Beer}
  }
\end{figure}

First, we compare the plots of the estimated MC and $K$ in Figure \ref{Fig: MC Beer}.
The results of BIC and NML are illustrated as an example.
In any method, the score was high at the end and beginning of the year,
reflecting the increased activities in transactions.
However, because the critical changes in the clustering structure
and changes due to ineffective components were mixed,
the sequence of $K$ had a lot of change points and was difficult to interpret their meanings.
On the other hand, MC identified the clustering structure
by discounting the effects of the ineffective components.
As a result, the sequence of MC highlighted the significant changes
at the end and beginning of the year.
It is also worthwhile noting that the differences of the scores
between the model selection methods were much smaller in MC than those in $K$;
it indicates that both BIC and NML estimates the similar clustering structure under the concept of MC
even though the number of components differs significantly.

\begin{figure}[htbp]
  \centering
  \subfigure[MC(total)]{
    \includegraphics[width=0.45\columnwidth]{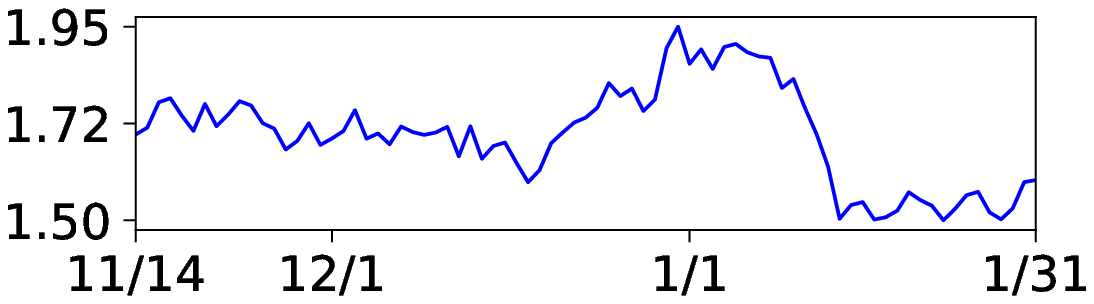}
  }
  \subfigure[MC(interaction)]{
    \includegraphics[width=0.45\columnwidth]{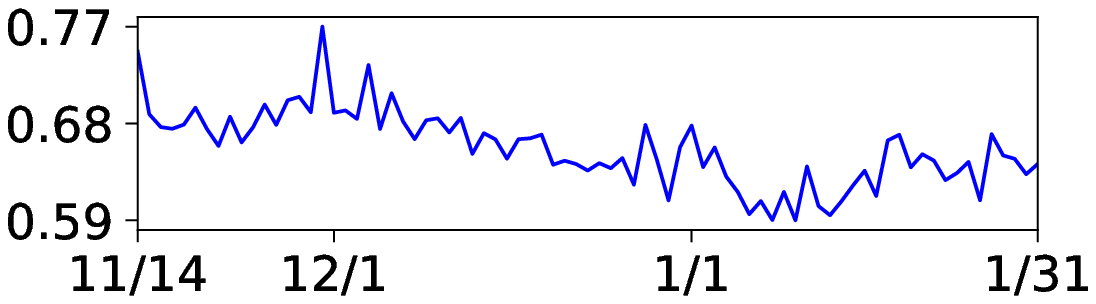}
  }
  \subfigure[Contribution(component 1)]{
    \includegraphics[width=0.45\columnwidth]{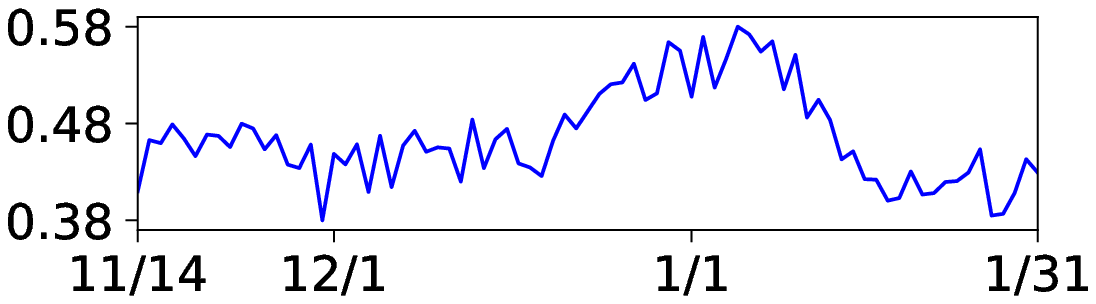}
  }
  \subfigure[Contribution(component 2)]{
    \includegraphics[width=0.45\columnwidth]{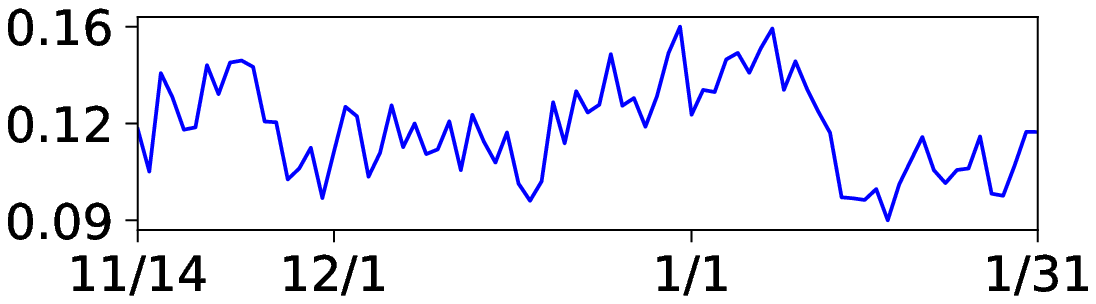}
  }
  \subfigure[Contribution(component 3)]{
    \includegraphics[width=0.45\columnwidth]{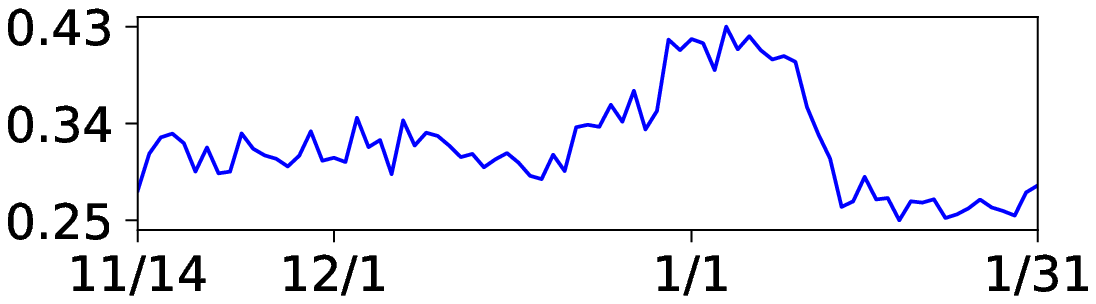}
  }
  \subfigure[Contribution(component 4)]{
    \includegraphics[width=0.45\columnwidth]{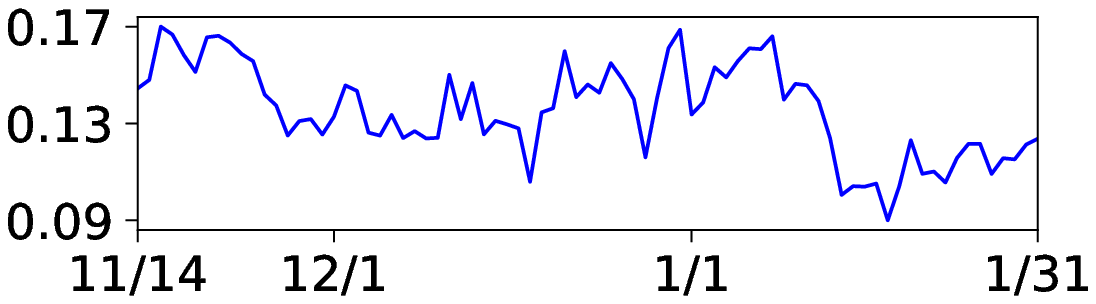}
  }
  \subfigure[W(component 1)]{
    \includegraphics[width=0.45\columnwidth]{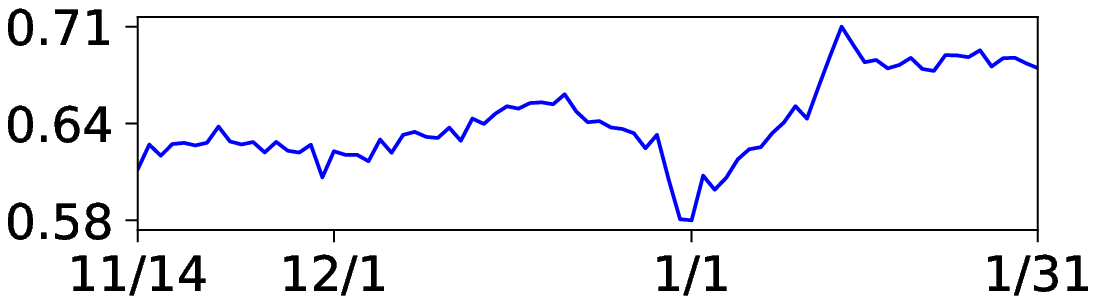}
  }
  \subfigure[W(component 2)]{
    \includegraphics[width=0.45\columnwidth]{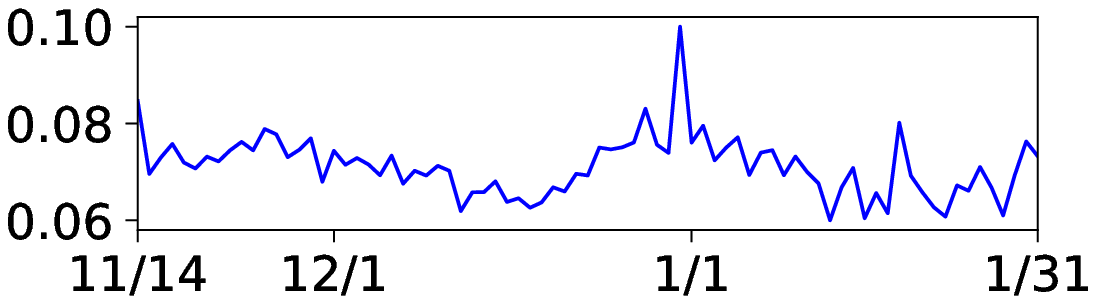}
  }
  \subfigure[W(component 3)]{
    \includegraphics[width=0.45\columnwidth]{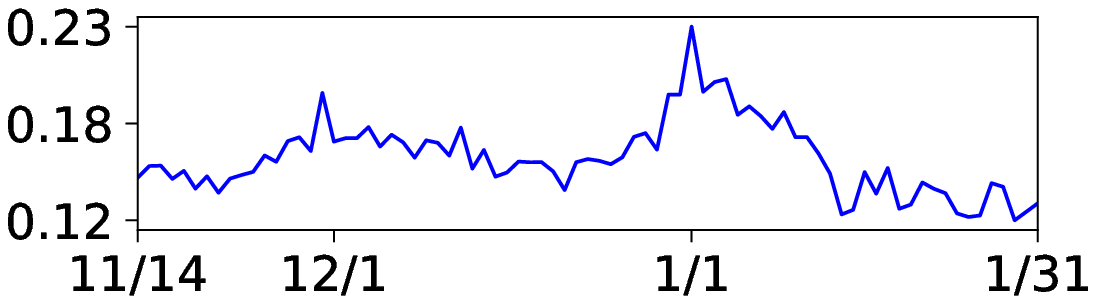}
  }
  \subfigure[W(component 4)]{
    \includegraphics[width=0.45\columnwidth]{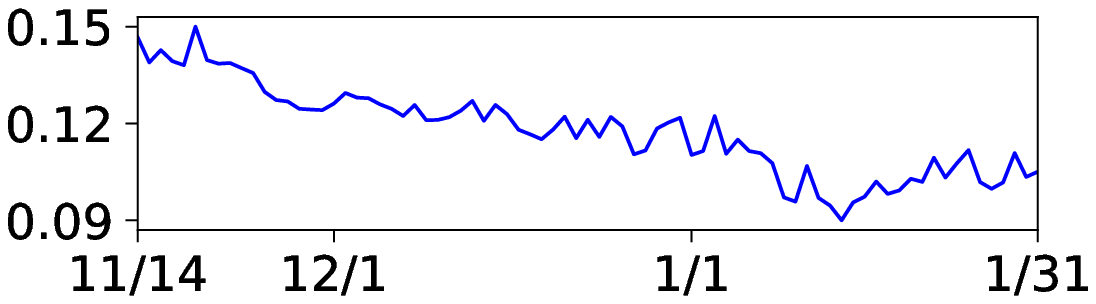}
  }
  \subfigure[MC(component 1)]{
    \includegraphics[width=0.45\columnwidth]{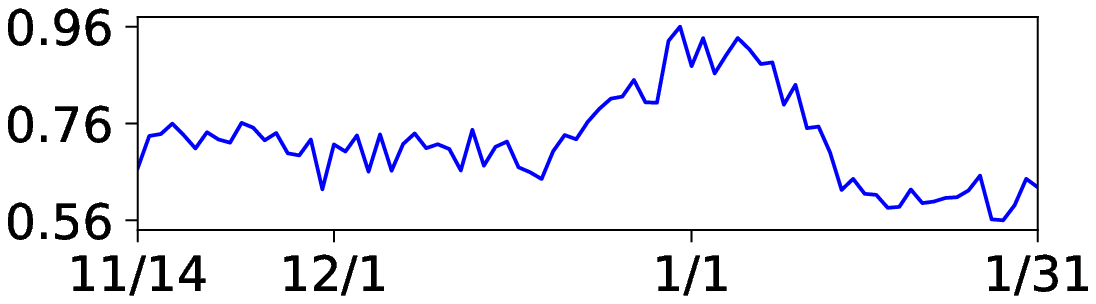}
  }
  \subfigure[MC(component 2)]{
    \includegraphics[width=0.45\columnwidth]{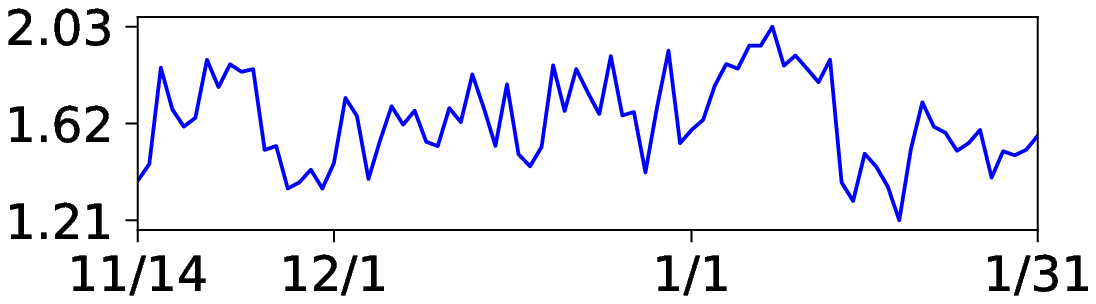}
  }
  \subfigure[MC(component 3)]{
    \includegraphics[width=0.45\columnwidth]{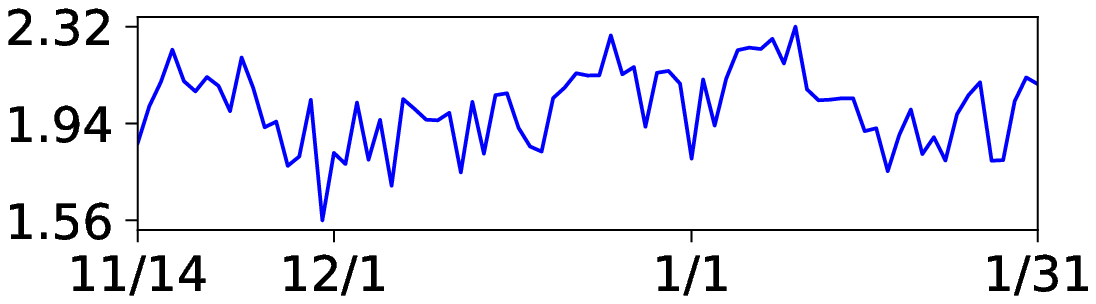}
  }
  \subfigure[MC(component 4)]{
    \includegraphics[width=0.45\columnwidth]{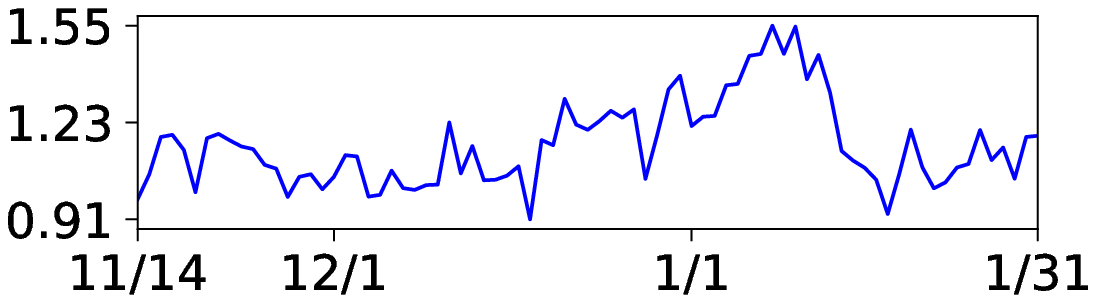}
  }
  \caption{
    Plots of the decomposition of MC with BIC in the beer Dataset.
    \label{Fig: decomposition Beer BIC}
  }
\end{figure}

\begin{figure}[htbp]
  \centering
  \subfigure[MC(total)]{
    \includegraphics[width=0.45\columnwidth]{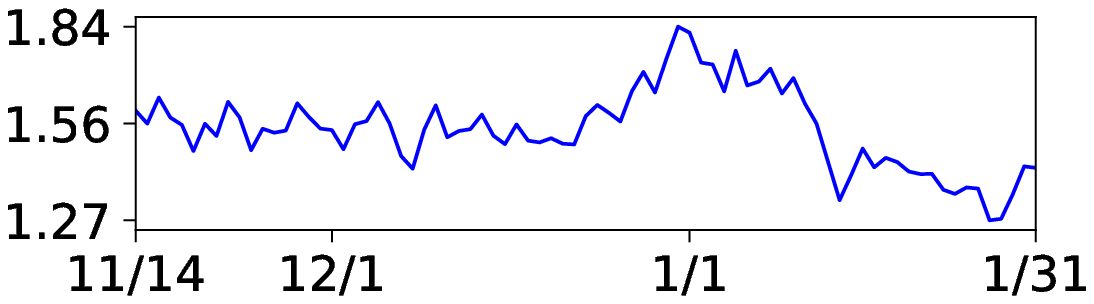}
  }
  \subfigure[MC(interaction)]{
    \includegraphics[width=0.45\columnwidth]{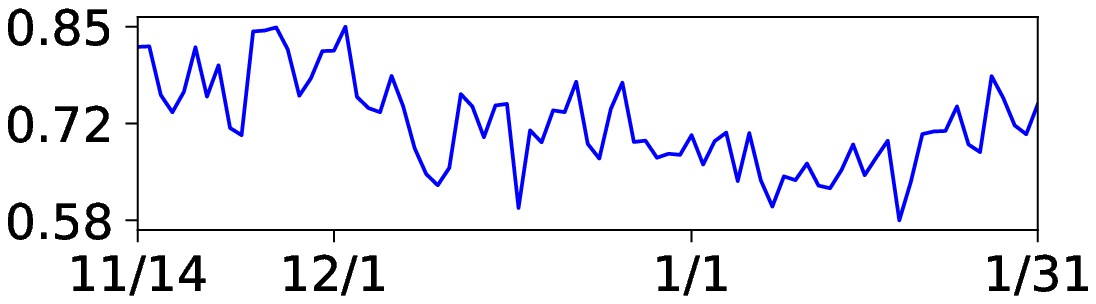}
  }
  \subfigure[Contribution(component 1)]{
    \includegraphics[width=0.45\columnwidth]{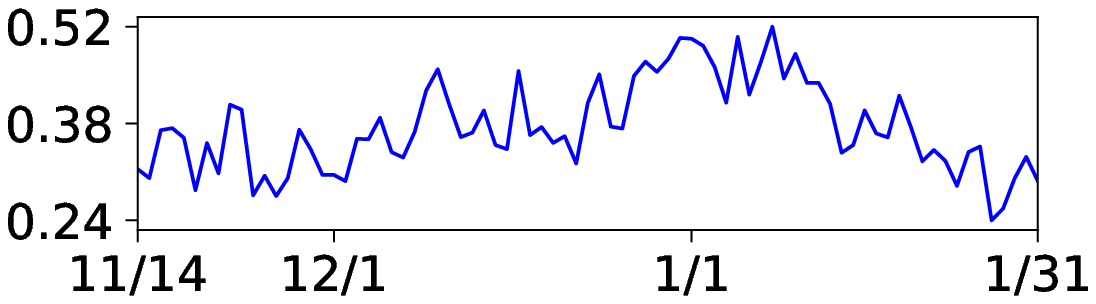}
  }
  \subfigure[Contribution(component 2)]{
    \includegraphics[width=0.45\columnwidth]{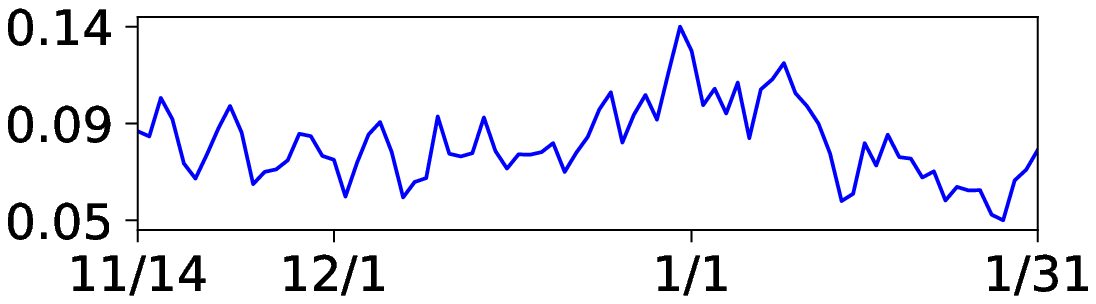}
  }
  \subfigure[Contribution(component 3)]{
    \includegraphics[width=0.45\columnwidth]{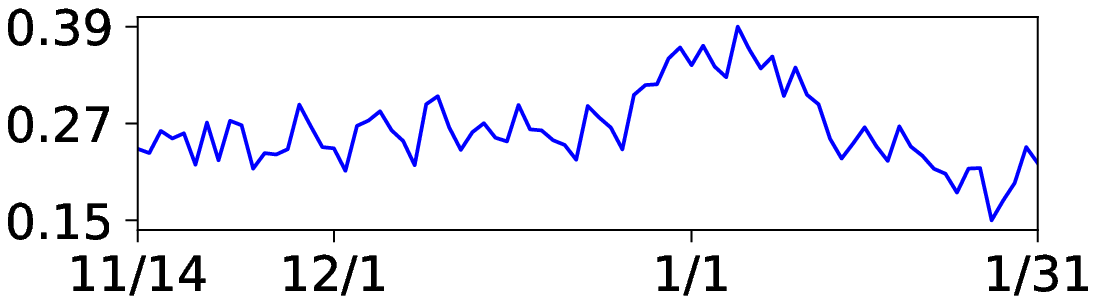}
  }
  \subfigure[Contribution(component 4)]{
    \includegraphics[width=0.45\columnwidth]{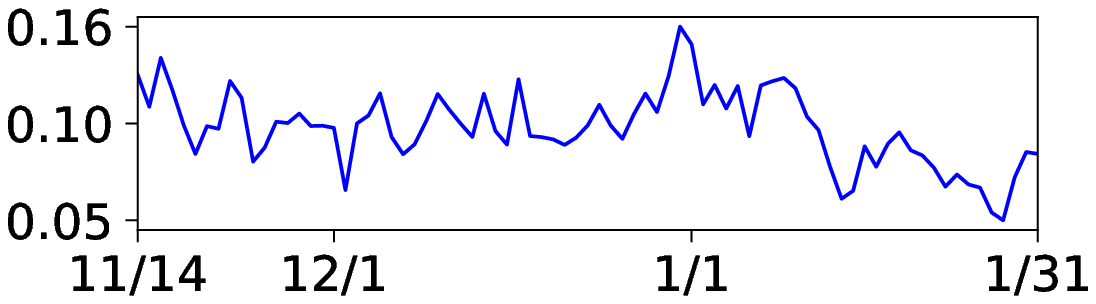}
  }
  \subfigure[W(component 1)]{
    \includegraphics[width=0.45\columnwidth]{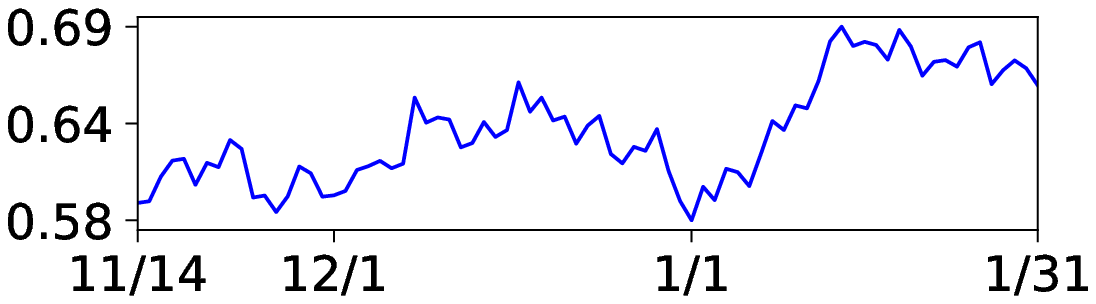}
  }
  \subfigure[W(component 2)]{
    \includegraphics[width=0.45\columnwidth]{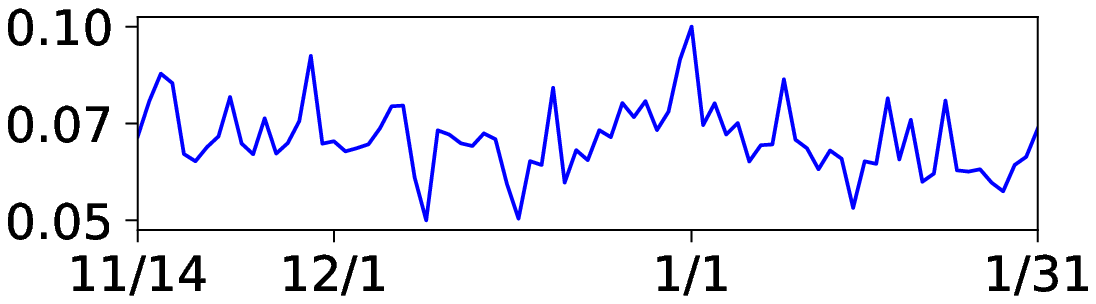}
  }
  \subfigure[W(component 3)]{
    \includegraphics[width=0.45\columnwidth]{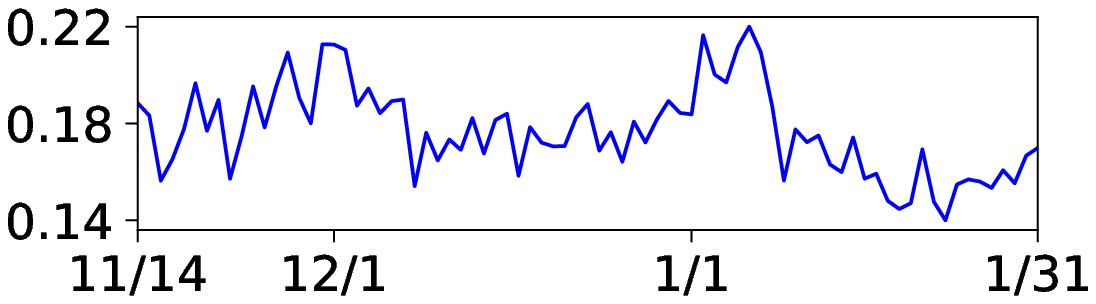}
  }
  \subfigure[W(component 4)]{
    \includegraphics[width=0.45\columnwidth]{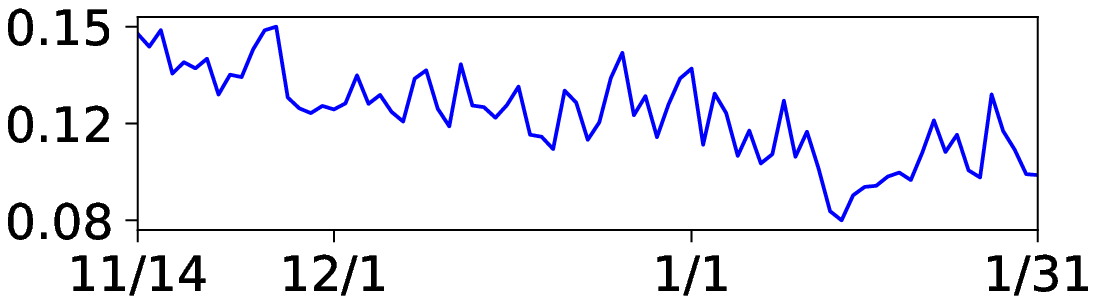}
  }
  \subfigure[MC(component 1)]{
    \includegraphics[width=0.45\columnwidth]{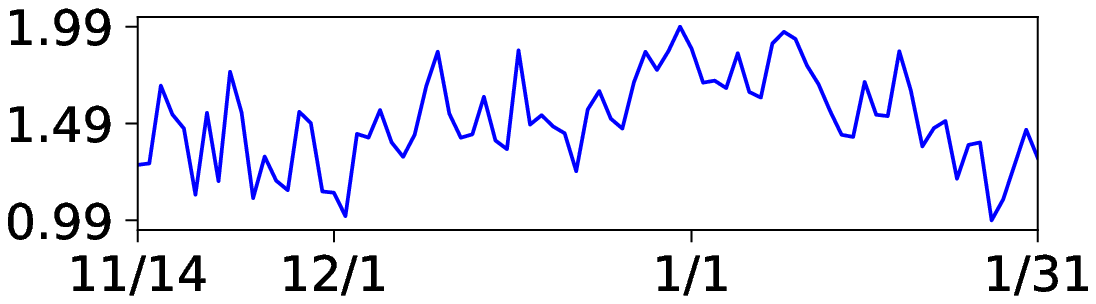}
  }
  \subfigure[MC(component 2)]{
    \includegraphics[width=0.45\columnwidth]{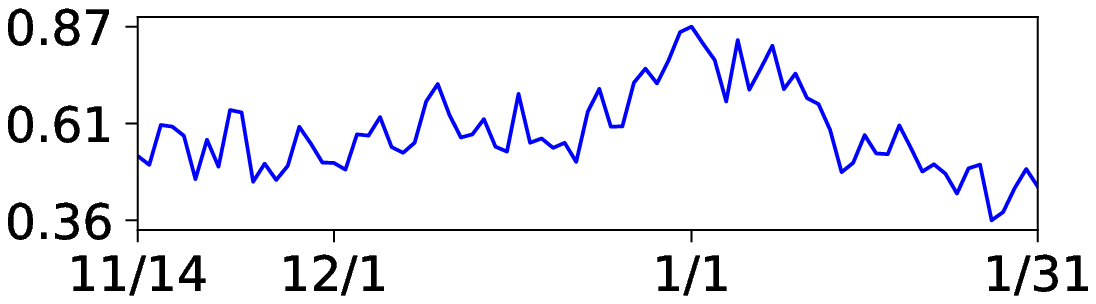}
  }
  \subfigure[MC(component 3)]{
    \includegraphics[width=0.45\columnwidth]{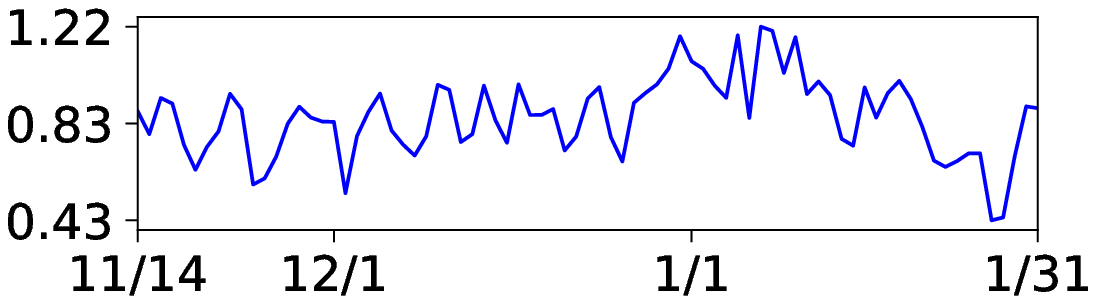}
  }
  \subfigure[MC(component 4)]{
    \includegraphics[width=0.45\columnwidth]{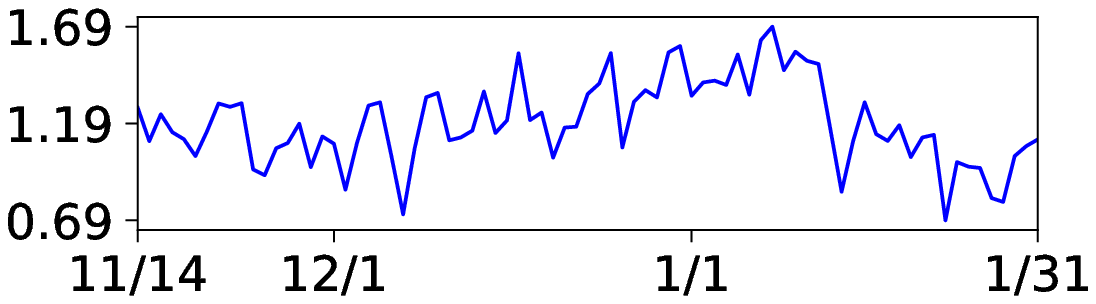}
  }
  \caption{
    Plots of the decomposition of MC with NML in the beer Dataset.
    \label{Fig: decomposition Beer NML}
  }
\end{figure}

Next, we discuss the results of the decomposition of MC.
We present the results of BIC and NML with $L = 4$ and $m = 1.5$.
The centers of the upper components are listed
in Table \ref{Tab: decomposition Beer BIC}
and Table \ref{Tab: decomposition Beer NML}, respectively,
and the plots of each decomposed value are illustrated
in Figure \ref{Fig: decomposition Beer BIC}
and Figure \ref{Fig: decomposition Beer NML}, respectively.
The indices of the upper components are manually rearranged
so that they correspond with each other;
then, it can be observed that the results were also similar to each other.
The structures can be extensively evaluated by analyzing the decomposed values.
For instance, let us analyze the decomposed values at the end and beginning of the year.
As evident from the tables, they had different characteristics.
It can be observed from the figures that the contributions increased in all components,
indicating that they were related to the increase in MC(total).
The weight of the component decreased in cluster 1 and increased in component 2 and 3,
indicating that the customers moved from component 1 to component 2 and 3.
Also, MC(component~$l$) increased in all components,
indicating that the complexity or diversity increased within them.

\begin{table}[htbp]
  \centering
  \begin{tabular}{r|c|c|c|c} \hline
                    & component 1 & component 2   & component 3   & component 4   \\ \hline
    beer(A)         & 0.09        & 0.44          & \textbf{1.93} & 0.16          \\
    beer(B)         & 0.07        & 0.23          & \textbf{0.96} & 0.06          \\
    beer(C)         & 0.07        & 0.20          & \textbf{0.83} & 0.07          \\
    beer(D)         & 0.05        & 0.20          & \textbf{0.58} & 0.05          \\
    beer(E)         & 0.03        & 0.06          & \textbf{0.35} & 0.03          \\
    beer(F)         & 0.03        & 0.06          & \textbf{0.35} & 0.02          \\
    beer(other)     & 0.04        & 0.12          & \textbf{0.69} & 0.10          \\
    beerlike(A)     & 0.02        & \textbf{5.85} & 0.23          & 0.07          \\
    beerlike(B)     & 0.09        & 0.57          & \textbf{0.80} & 0.21          \\
    beerlike(C)     & 0.10        & 0.63          & \textbf{0.83} & 0.22          \\
    beerlike(D)     & 0.07        & 0.40          & \textbf{0.57} & 0.18          \\
    beerlike(E)     & 0.04        & 0.12          & \textbf{0.51} & 0.06          \\
    beerlike(F)     & 0.04        & 0.20          & \textbf{0.34} & 0.13          \\
    beerlike(G)     & 0.05        & 0.10          & \textbf{0.40} & 0.06          \\
    beerlike(H)     & 0.03        & 0.09          & \textbf{0.26} & 0.04          \\
    beerlike(other) & 0.09        & 1.27          & 1.11          & \textbf{6.78} \\ \hline
  \end{tabular}
  \caption{
    Centers of the upper components estimated by BIC in the beer dataset.
    For each dimension, the maximum value is denoted in boldface.
    \label{Tab: decomposition Beer BIC}
  }
\end{table}

\begin{table}[htbp]
  \centering
  \begin{tabular}{r|c|c|c|c} \hline
                    & component 1 & component 2   & component 3   & component 4   \\ \hline
    beer(A)         & 0.08        & 0.48          & \textbf{1.90} & 0.12          \\
    beer(B)         & 0.04        & 0.30          & \textbf{1.04} & 0.07          \\
    beer(C)         & 0.05        & 0.20          & \textbf{0.95} & 0.04          \\
    beer(D)         & 0.04        & 0.19          & \textbf{0.64} & 0.09          \\
    beer(E)         & 0.02        & 0.06          & \textbf{0.38} & 0.02          \\
    beer(F)         & 0.02        & 0.07          & \textbf{0.40} & 0.01          \\
    beer(other)     & 0.03        & 0.11          & \textbf{0.68} & 0.19          \\
    beerlike(A)     & 0.02        & \textbf{5.79} & 0.21          & 0.07          \\
    beerlike(B)     & 0.10        & 0.52          & \textbf{0.73} & 0.18          \\
    beerlike(C)     & 0.11        & 0.61          & \textbf{0.70} & 0.21          \\
    beerlike(D)     & 0.06        & 0.49          & \textbf{0.52} & 0.24          \\
    beerlike(E)     & 0.04        & 0.12          & \textbf{0.47} & 0.07          \\
    beerlike(F)     & 0.04        & 0.18          & \textbf{0.30} & 0.24          \\
    beerlike(G)     & 0.04        & 0.11          & \textbf{0.44} & 0.07          \\
    beerlike(H)     & 0.02        & 0.10          & \textbf{0.23} & 0.09          \\
    beerlike(other) & 0.08        & 1.42          & 1.08          & \textbf{6.54} \\ \hline
  \end{tabular}
  \caption{
    Centers of the upper components estimated by NML in the beer dataset.
    For each dimension, the maximum value is denoted in boldface.
    \label{Tab: decomposition Beer NML}
  }
\end{table}

\subsubsection{House dataset}

We discuss the results of the house dataset,
obtained from the UCI Machine Learning Repository \citep{DheehuCasey2017}.
The dataset comprises the records of electricity consumption
in a house every five minutes from December 16th, 2006 to November 26th, 2010.
The dataset $\mathcal{X}$ is constructed as follows.
The time unit is 15 minutes from 00:00-00:15 to 23:45-24:00.
For each $t$, the data $\{ x_{n, t} \}_{n = 1}^N$ denotes the set of the records
on the various days included in the $t$-th time unit.
The dimension $d$ of the vector is 3,
which corresponds to the metering of the following three points:
\begin{itemize}
  \item metering(A): a kitchen.
  \item metering(B): a laundry room.
  \item metering(C): a water-heater and an air-conditioner.
\end{itemize}

\begin{figure}[htbp]
  \centering
  \includegraphics[width=0.9\columnwidth]{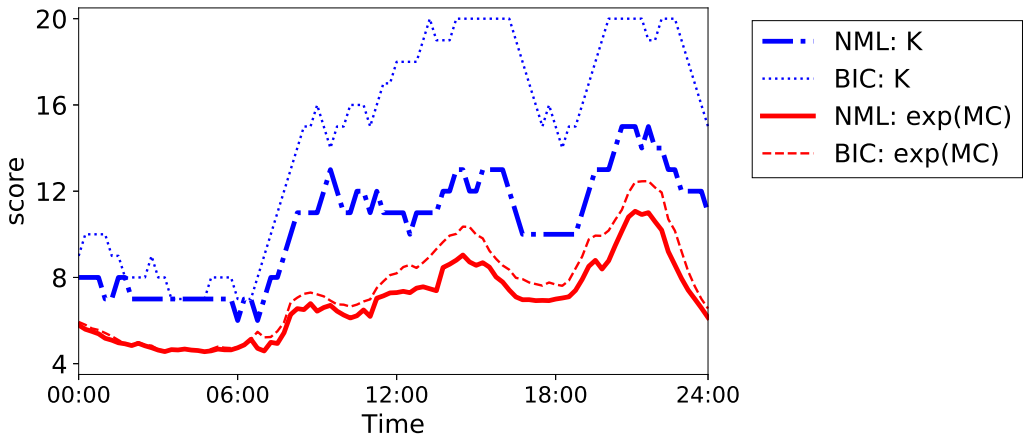}
  \caption{
    Plots of the sequences of the MC and $K$ in the house dataset.
    \label{Fig: MC House}
  }
\end{figure}

First, we compare the plots of the estimated $K$ and the corresponding MC in Figure \ref{Fig: MC House}.
The results of BIC and NML are illustrated as an example.
It can be observed from the figure
that the MC smoothly connected the discrete changes in $K$;
therefore, MC expressed gradual changes in the dataset more effectively than $K$.
Also, the MCs in BIC and NML were more similar each other than $K$ as well as in the beer dataset.
The values of MC started increasing from around 7:00 a.m.;
after slight fluctuations, the value reached its peak around 21:00.
Therefore, MC seemed to represent the amount of activities in this house.

\begin{figure}[htbp]
  \centering
  \subfigure[MC(total)]{
    \includegraphics[width=0.45\columnwidth]{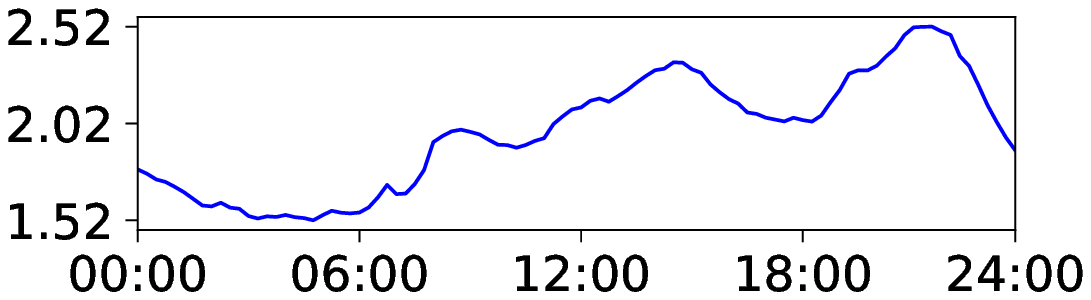}
  }
  \subfigure[MC(interaction)]{
    \includegraphics[width=0.45\columnwidth]{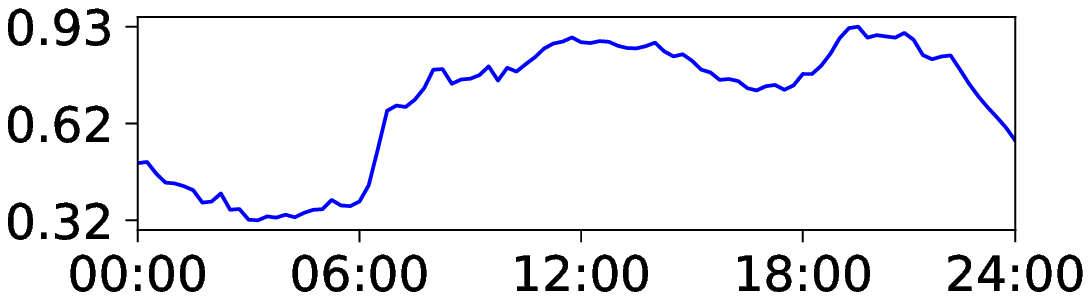}
  }
  \subfigure[Contribution(component 1)]{
    \includegraphics[width=0.45\columnwidth]{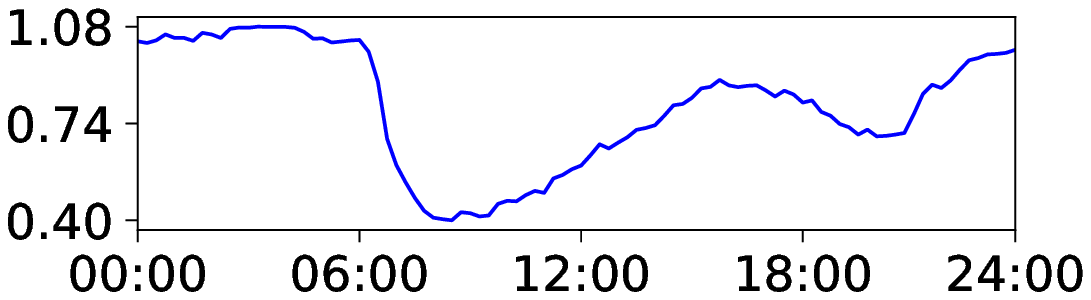}
  }
  \subfigure[Contribution(component 2)]{
    \includegraphics[width=0.45\columnwidth]{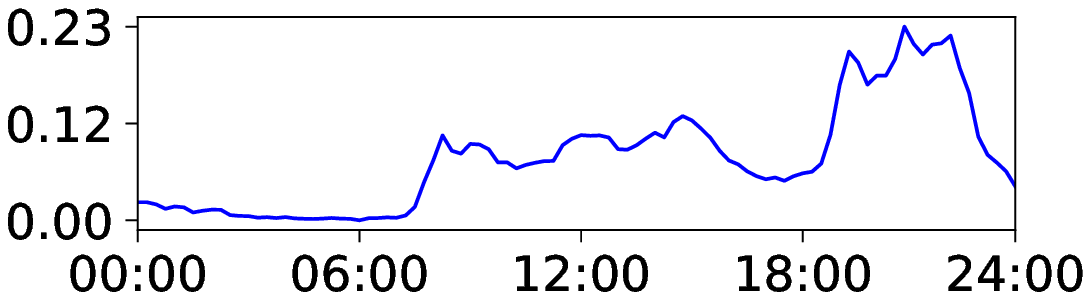}
  }
  \subfigure[Contribution(component 3)]{
    \includegraphics[width=0.45\columnwidth]{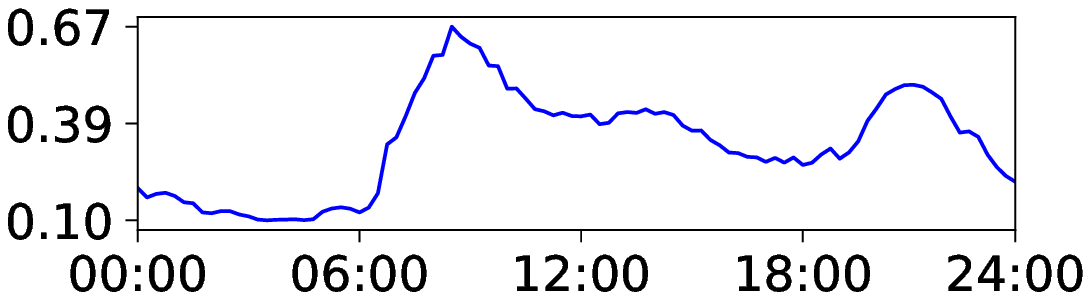}
  }
  \subfigure[Contribution(component 4)]{
    \includegraphics[width=0.45\columnwidth]{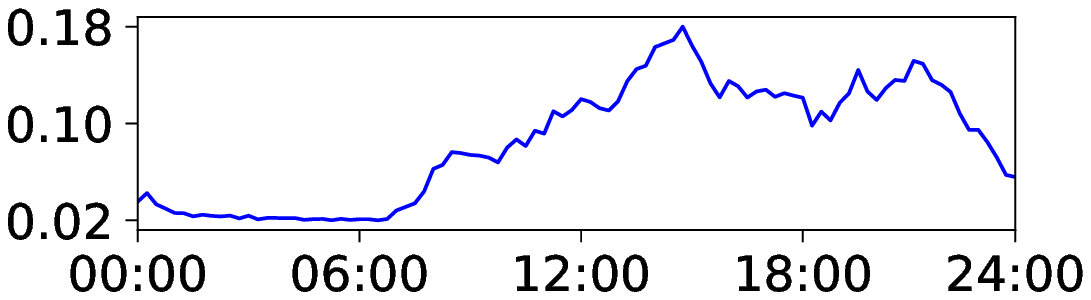}
  }
  \subfigure[W(component 1)]{
    \includegraphics[width=0.45\columnwidth]{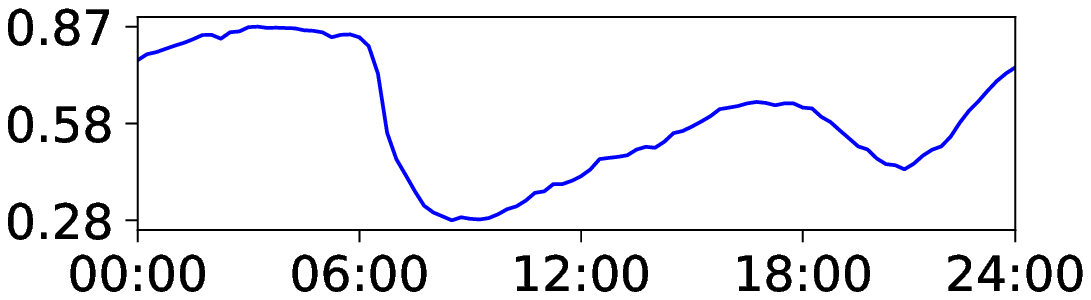}
  }
  \subfigure[W(component 2)]{
    \includegraphics[width=0.45\columnwidth]{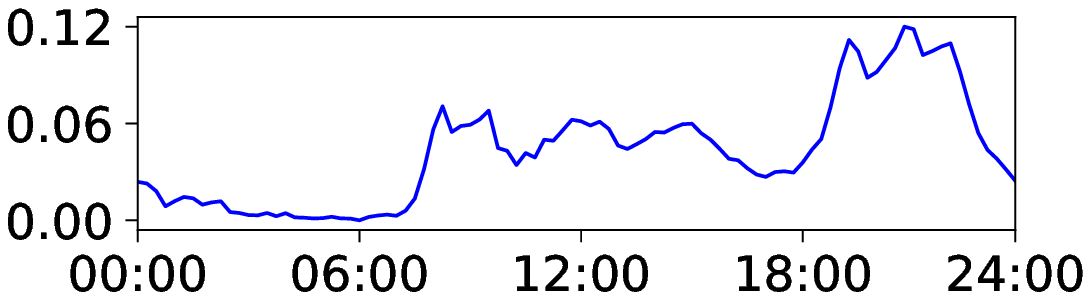}
  }
  \subfigure[W(component 3)]{
    \includegraphics[width=0.45\columnwidth]{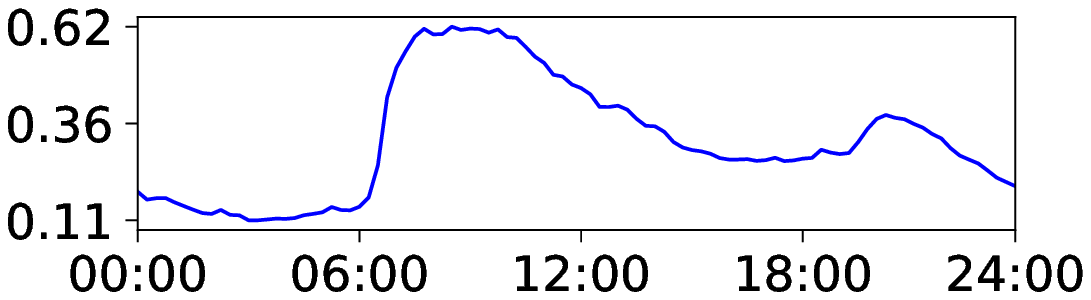}
  }
  \subfigure[W(component 4)]{
    \includegraphics[width=0.45\columnwidth]{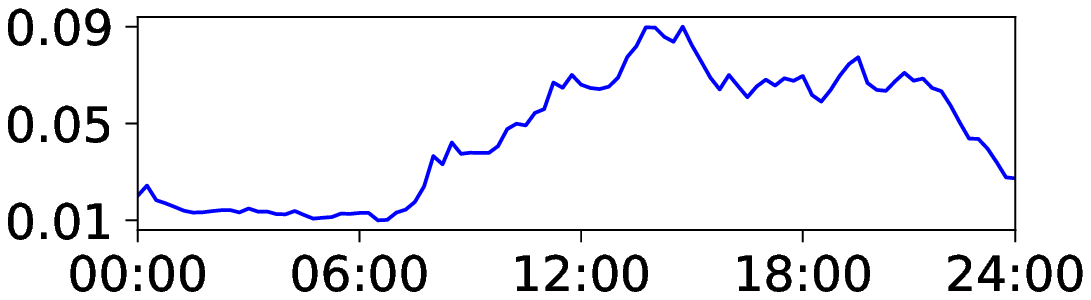}
  }
  \subfigure[MC(component 1)]{
    \includegraphics[width=0.45\columnwidth]{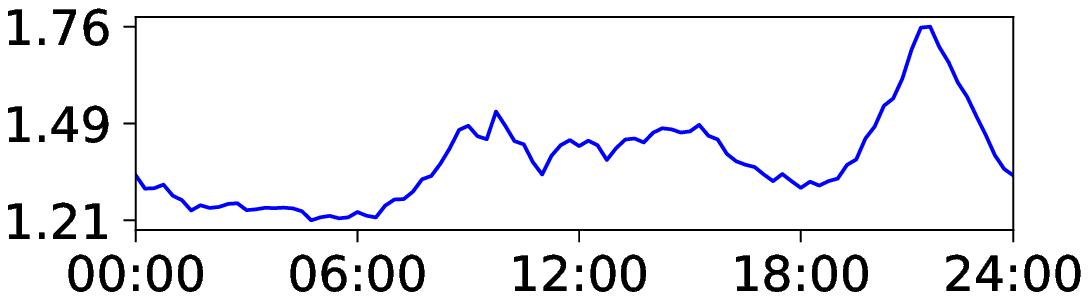}
  }
  \subfigure[MC(component 2)]{
    \includegraphics[width=0.45\columnwidth]{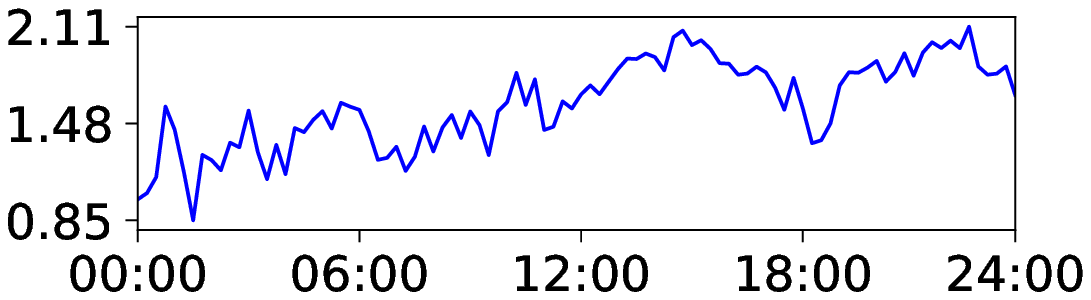}
  }
  \subfigure[MC(component 3)]{
    \includegraphics[width=0.45\columnwidth]{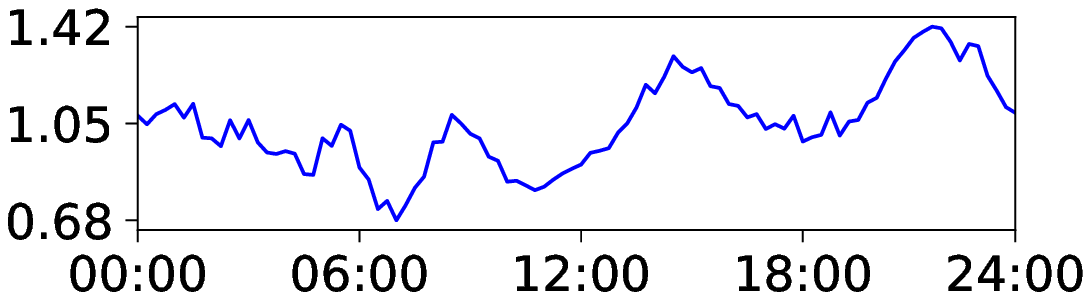}
  }
  \subfigure[MC(component 4)]{
    \includegraphics[width=0.45\columnwidth]{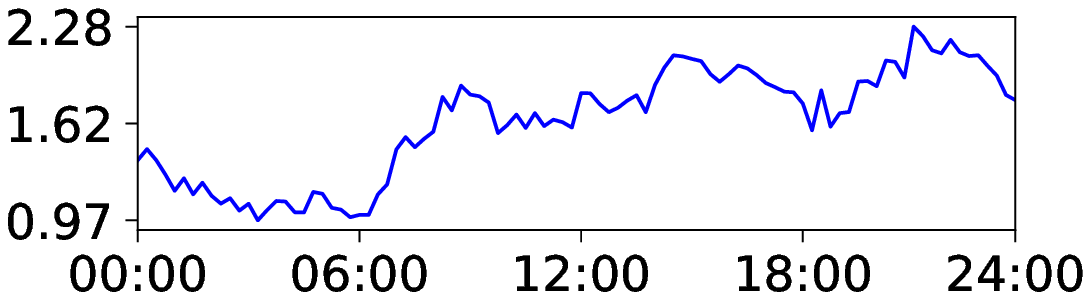}
  }
  \caption{
    Plots of the decomposition of MC with BIC in the house Dataset.
    \label{Fig: decomposition House BIC}
  }
\end{figure}

\begin{figure}[htbp]
  \centering
  \subfigure[MC(total)]{
    \includegraphics[width=0.45\columnwidth]{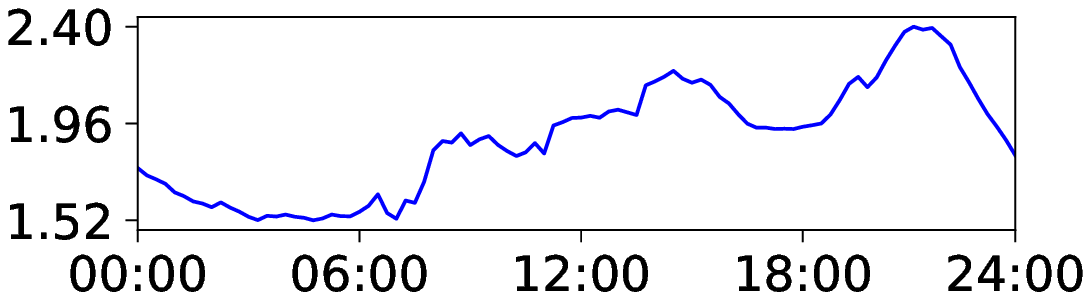}
  }
  \subfigure[MC(interaction)]{
    \includegraphics[width=0.45\columnwidth]{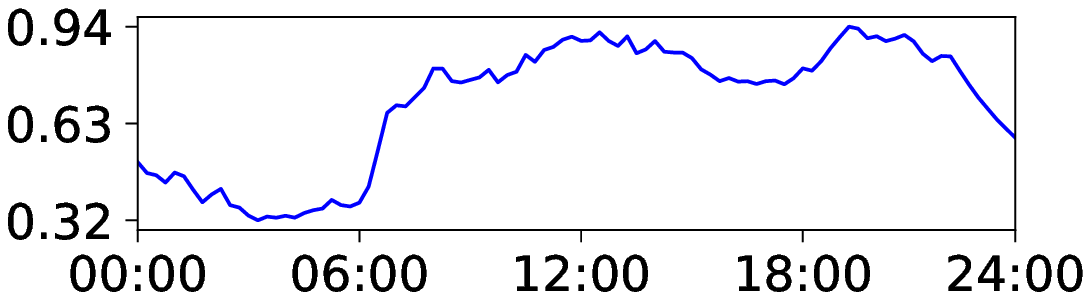}
  }
  \subfigure[Contribution(component 1)]{
    \includegraphics[width=0.45\columnwidth]{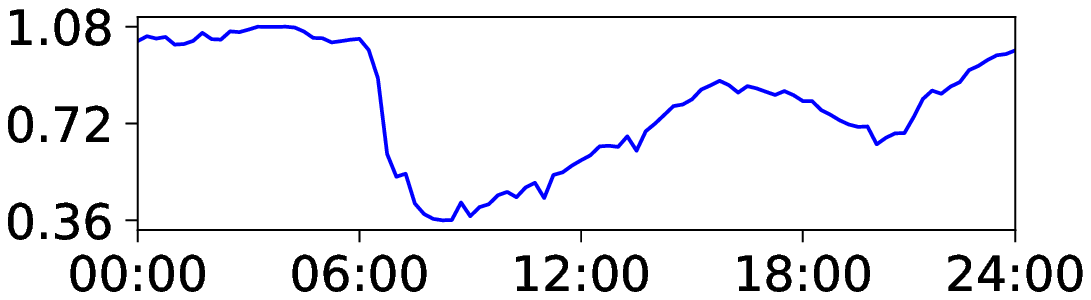}
  }
  \subfigure[Contribution(component 2)]{
    \includegraphics[width=0.45\columnwidth]{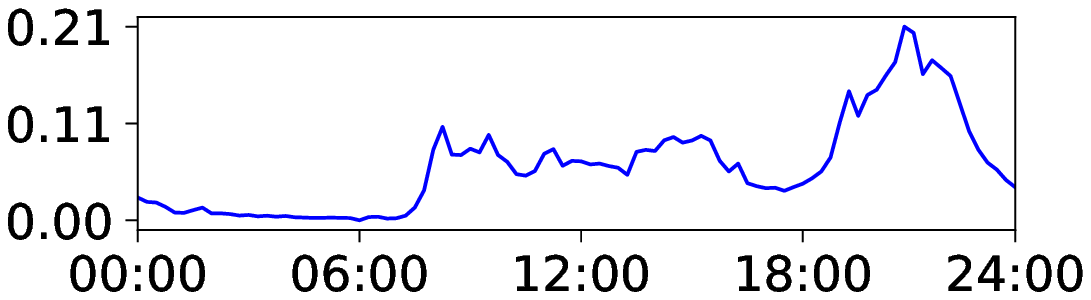}
  }
  \subfigure[Contribution(component 3)]{
    \includegraphics[width=0.45\columnwidth]{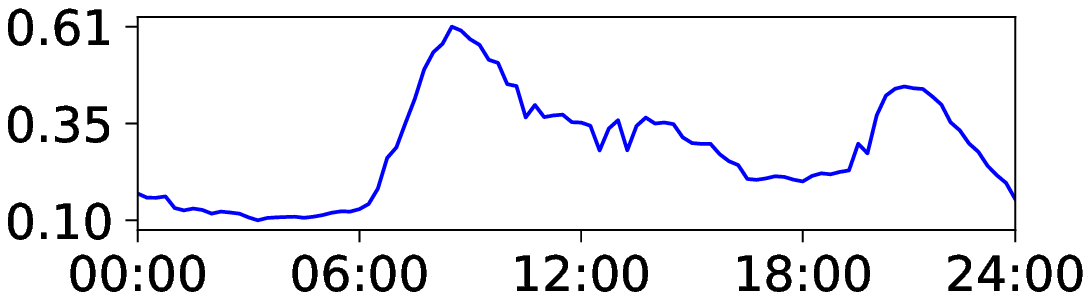}
  }
  \subfigure[Contribution(component 4)]{
    \includegraphics[width=0.45\columnwidth]{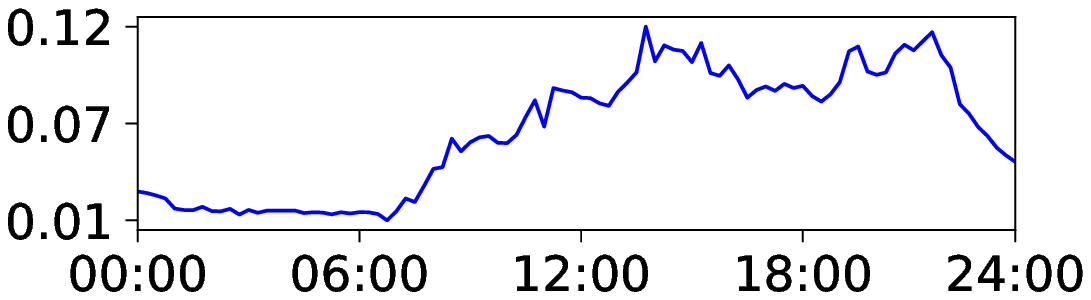}
  }
  \subfigure[W(component 1)]{
    \includegraphics[width=0.45\columnwidth]{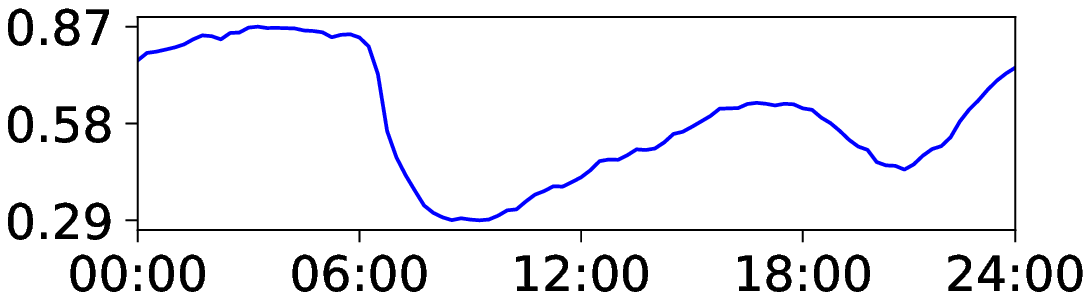}
  }
  \subfigure[W(component 2)]{
    \includegraphics[width=0.45\columnwidth]{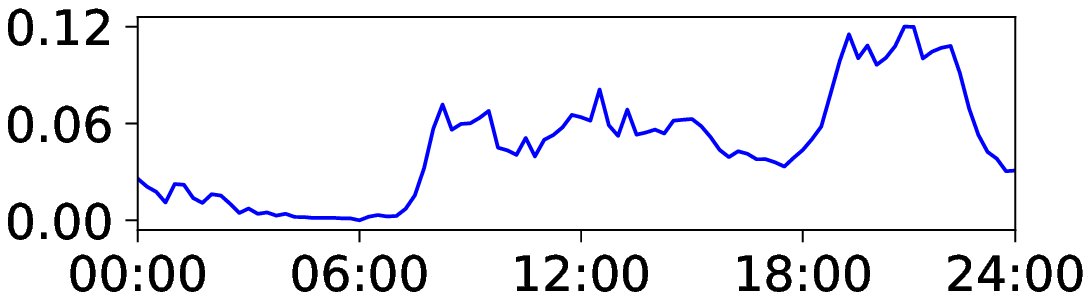}
  }
  \subfigure[W(component 3)]{
    \includegraphics[width=0.45\columnwidth]{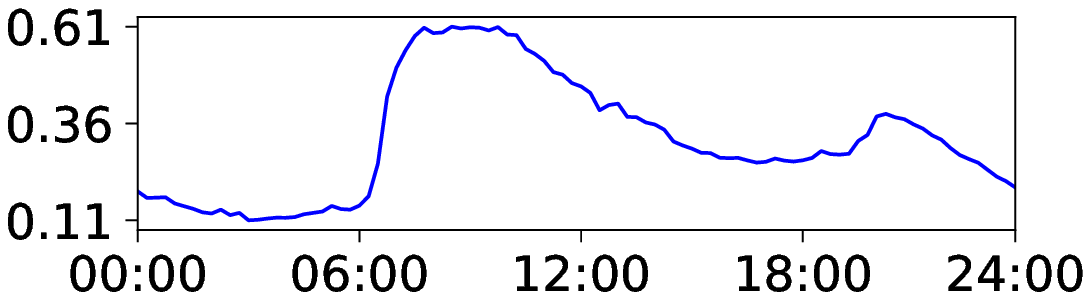}
  }
  \subfigure[W(component 4)]{
    \includegraphics[width=0.45\columnwidth]{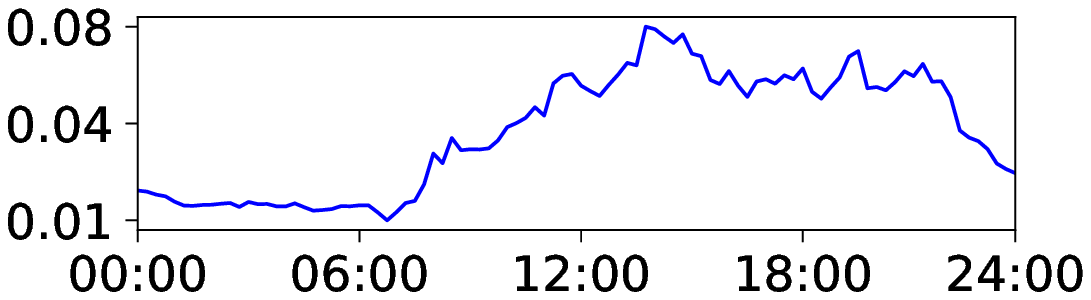}
  }
  \subfigure[MC(component 1)]{
    \includegraphics[width=0.45\columnwidth]{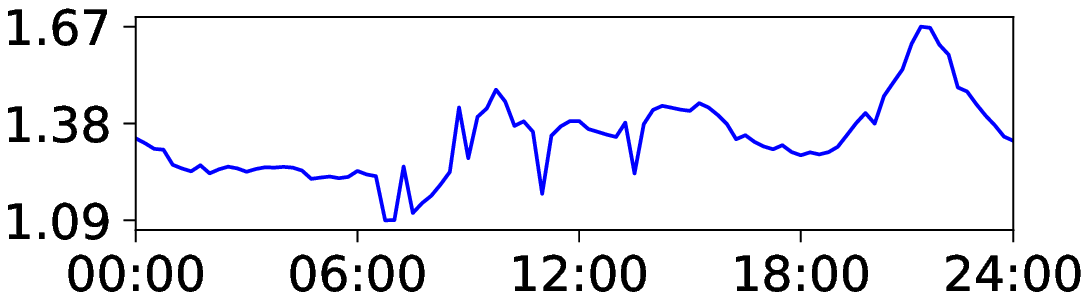}
  }
  \subfigure[MC(component 2)]{
    \includegraphics[width=0.45\columnwidth]{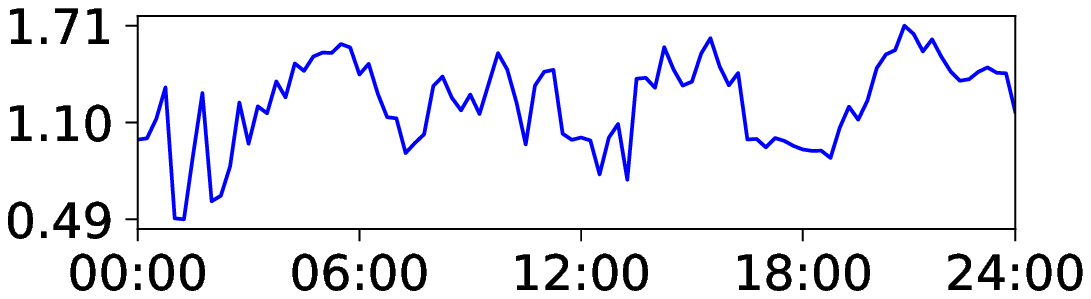}
  }
  \subfigure[MC(component 3)]{
    \includegraphics[width=0.45\columnwidth]{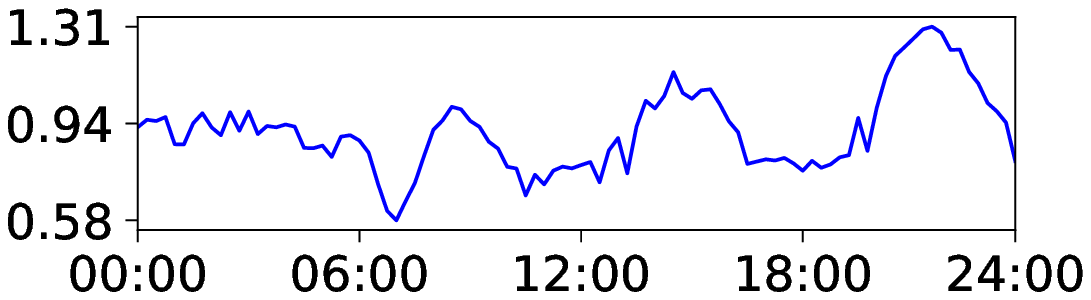}
  }
  \subfigure[MC(component 4)]{
    \includegraphics[width=0.45\columnwidth]{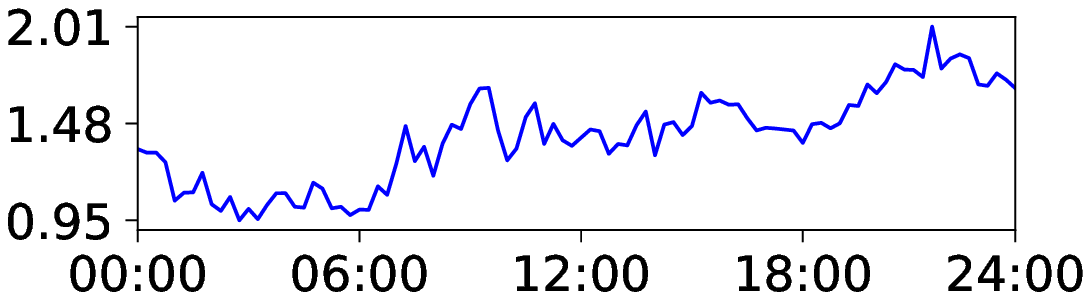}
  }
  \caption{
    Plots of the decomposition of MC with NML in the house Dataset.
    \label{Fig: decomposition House NML}
  }
\end{figure}

Next, we discuss the results of the decomposition of MC.
We present the results of BIC and NML with $L = 4$ and $m = 1.5$.
The centers of the upper components are listed
in Table \ref{Tab: decomposition House BIC}
and Table \ref{Tab: decomposition House NML}, respectively,
and the plots of each decomposed value are illustrated
in Figure \ref{Fig: decomposition House BIC}
and Figure \ref{Fig: decomposition House NML}, respectively.
The indices of the upper components are manually rearranged
so that they correspond with each other;
then, it can be observed that the results were also similar to each other.
The structures can be extensively evaluated by analyzing the decomposed values.
For instance, let us analyze the decomposed values in component 3.
It can be observed from the tables
that the value in metering(C) was specifically high in this component.
Looking at contribution(component 3),
there were two peaks around 9:00 and 21:00; it represented the increased activities in this component.
However, the proportions of the weight and MC were different.
W(component 3) was specifically high at 9:00, indicating that
the first half of the peaks was due to the increase in the weight of the component;
whereas, MC(component 3) was specifically high at 21:00,
indicating that the second half of the peaks
was due to the increase in the complexity within the component.

\begin{table}[htbp]
  \centering
  \begin{tabular}{c|c|c|c|c}
    \hline
                & component 1 & component 2   & component 3   & component 4   \\ \hline
    metering(A) & 0.04        & \textbf{4.47} & 0.13          & 0.41          \\
    metering(B) & 0.53        & 0.89          & 0.56          & \textbf{4.40} \\
    metering(C) & 0.75        & 3.34          & \textbf{4.37} & 2.96          \\ \hline
  \end{tabular}
  \caption{
    Centers of the upper components estimated by BIC in the house dataset.
    For each dimension, the maximum value is denoted in boldface.
    \label{Tab: decomposition House BIC}
  }
\end{table}

\begin{table}[htbp]
  \centering
  \begin{tabular}{r|c|c|c|c} \hline
                & component 1 & component 2   & component 3   & component 4   \\ \hline
    metering(A) & 0.04        & \textbf{4.24} & 0.11          & 0.35          \\
    metering(B) & 0.53        & 1.00          & 0.57          & \textbf{4.48} \\
    metering(C) & 0.76        & 3.37          & \textbf{4.38} & 2.93          \\ \hline
  \end{tabular}
  \caption{
    Centers of the upper components estimated by NML in the house dataset.
    For each dimension, the maximum value is denoted in boldface.
    \label{Tab: decomposition House NML}
  }
\end{table}

\section{Conclusion \label{Sec: conclusion}}

We proposed the concept of MC to measure the cluster size continuously in the mixture model.
We first pointed out that the cluster size might not be equal to the mixture size
when the mixture model had overlap or weight bias;
then, we introduced MC as an extended concept of the cluster size considering the effects of them.
We also presented methods to decompose the MC according to the mixture hierarchies,
which helped us in extensively analyzing the substructures.
Subsequently, we implemented the MC and its decomposition
to the gradual clustering change detection problems.
We conducted experiments to verify that the MC effectively elucidate the clustering changes.
In the artificial data experiments, MC found the clustering changes significantly earlier
in the case where the overlap or weight bias was correctly estimated.
In the real data experiments, MC expressed the gradual changes better than $K$
because it discerned the significant and insignificant changes
and smoothly connected the discrete changes in $K$.
We also found that the MC took similar values for each model selection method;
it indicates that the estimated clustering structures are alike under the concept of MC.
Moreover, its decomposition enabled us to evaluate the contents of changes.

Issues of the MC will be tackled in future study.
For example, it does not capture the clustering structure well
when the number of the components is underestimated;
thus, we need to explore the model selection methods that are more compatible with MC.
Also, we further need to study its theoretical aspects,
such as convergence and methods for approximating the mutual information.

\section*{Acknowledgement}

The marketing dataset was provided by Hakuhodo, Inc. and M-CUBE, inc.
This work was partially supported by JST KAKENHI 191400000190 and JST-AIP JPMJCR19U4.

\appendix

\section{Proofs \label{Sec: proofs}}

In this section, we describe the proofs of the propositions and theorem stated in the text.

\subsection{Proof of Proposition \ref{Prop: entirely overlap}}

\begin{proof}
  If $g_1 = \dots = g_K$, then
  \[
    \mathrm{MC}(\{ \pi_k, g_k \}_{k = 1}^K; \{ x_n \}_{n = 1}^N)
    = \frac{1}{N} \sum_{n = 1}^N \sum_{k = 1}^K \frac{\pi_k g_k(x_n)}{f(x_n)} \log 1
    = 0 \eqspace .
  \]
\end{proof}

\subsection{Proof of Proposition \ref{Prop: entirely separated}}

\begin{proof}
  We see from the assumption that
  \[
    \frac{\pi_k g_k(x_n)}{f(x_n)} \log \left( \frac{g_k(x_n)}{f(x_n)} \right) =
    \begin{cases}
      - \log \pi_k & (g_k(x_n) > 0) \eqspace , \\
      0            & (g_k(x_n) = 0) \eqspace ,
    \end{cases}
  \]
  where $0 \log 0 \coloneqq 0$. Note that there are no problems even when $\pi_k = 0$
  because $\#\{ x_n \mid g_k(x_n) > 0 \} = 0$ in this case.
  Then, we can calculate the MC as
  \[
    \mathrm{MC}(\{ \pi_k, g_k \}_{k = 1}^K; \{ x_n \}_{n = 1}^N)
    = - \frac{1}{N} \sum_{k = 1}^K \#\{ x_n \mid g_k(x_n) > 0 \} \log \pi_k
    = H(Z) \eqspace .
  \]
\end{proof}

\subsection{Proof of Proposition \ref{Prop: bounds}}

\begin{proof}
  For all $\{ \pi_k \}_{k = 1}^K$, we derive
  $\mathrm{MC}(\{ \pi_k, g_k \}_{k = 1}^K; \{ x_n \}_{n = 1}^N) \geq 0$ as
  \begin{align*}
    \mathrm{MC}(\{ \pi_k, g_k \}_{k = 1}^K; \{ x_n \}_{n = 1}^N)
     & = \frac{1}{N} \sum_{n = 1}^N \sum_{k = 1}^K
    \frac{\pi_k g_k(x_n)}{f(x_n)}
    \log \left( \frac{g_k(x_n)}{f(x_n)} \right)           \\
     & \geq \frac{1}{N} \sum_{n = 1}^N \sum_{k = 1}^K
    \frac{\pi_k g_k(x_n)}{f(x_n)}
    \left( 1 - \frac{f(x_n)}{g_k(x_n)} \right)            \\
     & = 0 \eqspace .
  \end{align*}
  Moreover, we see from the Karush-Kuhn-Tucker condition that
  \[
    \pi_k = \frac{1}{N} \sum_{n = 1}^N \frac{\pi_k g_k(x_n)}{f(x_n)}.
  \]
  From this,
  we derive $\mathrm{MC}(\{ \pi_k, g_k \}_{k = 1}^K; \{ x_n \}_{n = 1}^N) \leq \log K$ as
  \begin{align*}
         & \ \mathrm{MC}(\{ \pi_k, g_k \}_{k = 1}^K; \{ x_n \}_{n = 1}^N) \\
    =    & \ \frac{1}{N} \sum_{n = 1}^N \sum_{k = 1}^K
    \frac{\pi_k g_k (x_n)}{f(x_n)}
    \log \left( \frac{g_k(x_n)}{f(x_n)} \right)                           \\
    =    & \ - \frac{1}{N} \sum_{n = 1}^N \sum_{k = 1}^K
    \frac{\pi_k g_k(x_n)}{f(x_n)} \log \pi_k
    + \frac{1}{N} \sum_{n = 1}^N \sum_{k = 1}^K
    \frac{\pi_k g_k (x_n)}{f (x_n)}
    \log \left( \frac{\pi_k g_k(x_n)}{f(x_n)} \right)                     \\
    =    & \ - \sum_{k = 1}^K \pi_k \log \pi_k
    + \frac{1}{N} \sum_{n = 1}^N \sum_{k = 1}^K
    \frac{\pi_k g_k (x_n)}{f (x_n)}
    \log \left(\frac{\pi_k g_k(x_n)}{f(x_n)} \right)                      \\
    \leq & \ \log K + 0                                                   \\
    =    & \ \log K \eqspace .
  \end{align*}
\end{proof}

\subsection{Proof of Proposition \ref{Prop: invariance}}

\begin{proof}
  It follows from the fact that
  \[
    \sum_{k = 1}^K \pi_k g_k (\cdot)
    = \sum_{k' = 1}^{K'} \pi_{k'} g_{k'} (\cdot), \quad
    \sum_{k = 1}^K \pi_k g_k (\cdot) \log g_k (\cdot)
    = \sum_{k' = 1}^{K'} \pi_{k'} g_{k'} (\cdot) \log g_{k'} (\cdot) \eqspace .
  \]
\end{proof}

\subsection{Proof of Theorem \ref{Thm: decomposition}}

\begin{proof}
  We can directly calculate as
  \begin{align*}
      & \ \mathrm{MC}(\{ \pi_k, g_k \}_{k = 1}^K; \{ x_n \}_{n = 1}^N)
    - \mathrm{MC}(\{ \rho_l, h_l \}_{l = 1}^L; \{ x_n \}_{n = 1}^N)    \\
    = & \ \frac{1}{N} \sum_{n = 1}^N
    \left[
      \sum_{k = 1}^K \frac{\pi_k g_k(x_n)}{f(x_n)}
      \log \left( \frac{g_k(x_n)}{f(x_n)} \right)
      - \sum_{l = 1}^L \frac{\rho_l h_l(x_n)}{f(x_n)}
      \log \left( \frac{h_l(x_n)}{f(x_n)} \right)
    \right]                                                            \\
    = & \ \frac{1}{N} \sum_{n = 1}^N \sum_{l = 1}^L \sum_{k = 1}^K
    \frac{\pi_k Q^{(l)}_k g_k(x_n)}{f(x_n)}
    \left[
      \log \left( \frac{g_k(x_n)}{f(x_n)} \right)
      - \log \left( \frac{h_l(x_n)}{f(x_n)} \right)
    \right]                                                            \\
    = & \ \frac{1}{N} \sum_{n = 1}^N \sum_{l = 1}^L \sum_{k = 1}^K
    \frac{\pi_k Q^{(l)}_k g_k(x_n) \rho_l h_l(x_n)}{f(x_n) \rho_l h_l(x_n)}
    \log \left( \frac{g_k(x_n)}{h_l (x_n)} \right)                     \\
    = & \ \frac{1}{N} \sum_{n = 1}^N \sum_{l = 1}^L \sum_{k = 1}^K
    \frac{w^{(l)}_n \phi^{(l)}_l g_k(x_n)}{h_l(x_n)}
    \log \left( \frac{g_k(x_n)}{h_l(x_n)} \right)                      \\
    = & \ \sum_{l = 1}^L \frac{1}{N}
    \frac{\sum_{n' = 1}^N w^{(l)}_{n'}}{\sum_{n' = 1}^N w^{(l)}_{n'}}
    \sum_{n = 1}^N w^{(l)}_n \sum_{k = 1}^K
    \frac{\phi^{(l)}_l g_k(x_n)}{h_l(x_n)}
    \log \left( \frac{g_k(x_n)}{h_l(x_n)} \right)                      \\
    = & \ \sum_{l = 1}^L W_l \cdot
    \mathrm{MC}(\{ \phi^{(l)}_k, g_k \}_{k = 1}^K; \{ x_n, w^{(l)}_n \}_{n = 1}^N) \eqspace .
  \end{align*}
\end{proof}

\vskip 0.2in
\bibliography{mcbib}

\end{document}